\documentclass{article} % For LaTeX2e
\usepackage{iclr2025_conference,times}

\usepackage{amsmath,amsfonts,bm}

\def\eqref#1{equation~\ref{#1}}
\def\1{\bm{1}}

\DeclareMathAlphabet{\mathsfit}{\encodingdefault}{\sfdefault}{m}{sl}
\SetMathAlphabet{\mathsfit}{bold}{\encodingdefault}{\sfdefault}{bx}{n}

\usepackage{hyperref}
\usepackage{url}
\usepackage{booktabs}
\usepackage{enumitem}
\usepackage{algorithm}
\usepackage{algorithmic}
\usepackage{graphicx}
\usepackage{wrapfig}
\usepackage{multirow}
\usepackage{tabularx}
\usepackage{xcolor,colortbl}
\usepackage{longtable}
\usepackage{multicol,caption}
\usepackage{subfig}
\usepackage{bbm}

\definecolor{papercolor}{HTML}{0668E1}
\hypersetup{
    colorlinks,
    linkcolor={papercolor!75!black},
    citecolor={papercolor!75!black},
    urlcolor={papercolor!75!black}
}
\title{Varying Shades of Wrong: Aligning LLMs with Wrong Answers Only}

\author{Jihan Yao\thanks{equal contribution} \ \textsuperscript{1} \ \ \ \ \ Wenxuan Ding\footnotemark[1] \ \textsuperscript{2} \ \ \ \ \ Shangbin Feng\footnotemark[1] \ \textsuperscript{1} \ \ \ \ \ Lucy Lu Wang\textsuperscript{1}\textsuperscript{3} \ \ \ \ \ Yulia Tsvetkov\textsuperscript{1} \\
\textsuperscript{1}University of Washington \ \ \ \ \ \textsuperscript{2}The University of Texas at Austin \ \ \ \ \ \textsuperscript{3} Allen Institute for AI \\
\texttt{\{jihany2, shangbin\}@cs.washington.edu} \ \ \ \texttt{wenxuand@utexas.edu}
}

\iclrfinalcopy % Uncomment for camera-ready version, but NOT for submission.
\begin{document}

\maketitle

\begin{abstract}
%We investigate \emph{wrong-over-wrong alignment}: 
In the absence of abundant reliable annotations for challenging tasks and contexts, how can we expand the frontier of LLM capabilities with potentially wrong answers? We focus on two research questions: (1) \emph{Can LLMs generate reliable preferences among wrong options?} And if so, (2) \emph{Would alignment with such wrong-over-wrong preferences be helpful?} We employ methods based on self-consistency, token probabilities, and LLM-as-a-judge to elicit wrong-over-wrong preferences, and fine-tune language models with preference optimization approaches using these synthesized preferences. Extensive experiments with seven LLMs and eight datasets demonstrate that (1) LLMs \emph{do} have preliminary capability in distinguishing various shades of wrong, achieving up to 20.9\% higher performance than random guess; (2) Alignment with wrong-over-wrong preferences helps LLMs to produce less wrong and sometimes even outright correct answers, while overall improving model calibration. Code and data are publicly available at \href{https://github.com/yaojh18/Varying-Shades-of-Wrong}{https://github.com/yaojh18/Varying-Shades-of-Wrong}.
\end{abstract}

\section{Introduction}
Post-training with preference optimization, a.k.a.~\emph{alignment}, has become a crucial part of the development of large language models (LLMs) \citep{touvron2023llama}. From online alignment with PPO \citep{schulman2017proximal} and RLHF \citep{ouyang2022training} algorithms, to recent developments in offline alignment with DPO \citep{rafailov2023direct}, LLMs improve by learning nuanced distinctions and separability between answers: a correct answer is preferred over an incorrect answer in reasoning problems \citep{wang2023making}, or a response that adequately completes a user instruction is preferred over a failing one \citep{dubois2023alpacafarm}, etc. This response ``correctness'' in alignment procedures and implementations typically comes from datasets with ground truth answers \citep{wang2023making}, human annotators \citep{ouyang2022training}, or even the model itself \citep{lee2023rlaif}, which are employed to produce preference pairs and/or reward model training data.

\begin{figure}[!t]
    \centering
    \includegraphics[width=1\linewidth]{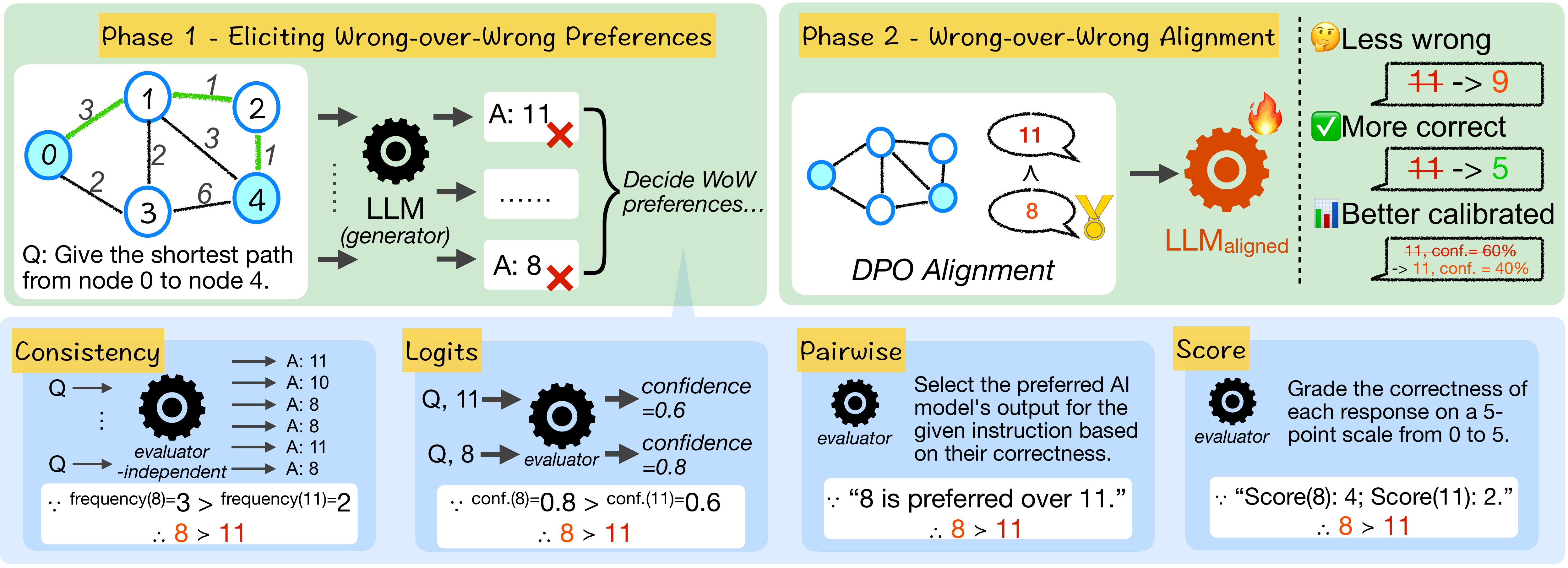}
    % \vspace{-0.1in}
    \caption{Two phases of aligning LLMs with wrong answers: \emph{eliciting wrong-over-wrong preferences} and \emph{wrong-over-wrong alignment}. In Phase 1, we employ four methods to elicit wrong-over-wrong preferences, based on answer consistency, logits-based confidence, and LLM-as-a-judge approaches. In Phase 2, we align LLMs with wrong-over-wrong preferences using DPO and expect to have less wrong, more correct, and better-calibrated answers.}
    \label{fig:intro}
    \vspace{-5pt}
\end{figure}

To overcome the bottleneck of relying on high-quality expert-annotated data, in this work, we ask: \emph{what if there is no correct answer in the alignment process}? What if the task comes without annotated ground-truths, is prohibitively expensive for human annotators, and even state-of-the-art models are too poor to consistently provide correct answers? We see an increasing number of challenging benchmarks such as theorem proving \citep{welleck2021naturalproofs, zhou2024refactor} and reasoning in low-resource languages \citep{huang2024mindmerger} where correct answers are simply unavailable. As a result, existing alignment procedures relying on ``right-over-wrong'' separability might struggle to expand the frontier of model capabilities. In response, we propose \emph{wrong-over-wrong alignment}, where LLMs evaluate the ``wrongness'' of incorrect answers and align by learning to prefer less-wrong answers over more-wrong ones in the absence of correct responses. We focus on tasks with clear right-wrong distinctions and present the outline of our work in Figure~\ref{fig:intro}:

RQ1: \emph{Can LLMs discriminate between varying shades of wrong and produce wrong-over-wrong preferences?} We employ four tasks where a reasonable ``proxy'' of wrongness exists, e.g., to find the shortest path in a network with a ground truth path length of 5, finding a path of length 8 is ``less wrong'' than a path of length 11 \citep{wang2023can}.\footnote{Wrongness proxies are by no means \emph{perfect}; they merely serve as objective and quantitative measures.} For each task, we sample multiple solutions to a problem, filter for incorrect answers, and employ LLMs to elicit wrong-over-wrong preferences via self-consistency \citep{wang2023selfconsistency}, token probabilities, pairwise comparison \citep{sun2024salmon}, and LLM-as-a-judge\citep{dubois2023alpacafarm}. We employ a quantitative proxy of wrongness in each task to help evaluate wrong-over-wrong preferences.

RQ2: \emph{Is LLM alignment with wrong-over-wrong preferences helpful?} We fine-tune language models to align them using the synthesized wrong-over-wrong preferences. We hypothesize that wrong-over-wrong alignment will make models better calibrated, learning a better representation space for correct and incorrect responses. The models might produce more correct answers as determined by ground truth\footnote{We employ datasets with ground-truths but only for evaluation proposes.}, or even if the answers are still incorrect, they will be less wrong based on the proxy.

Extensive experiments with seven open and proprietary models across eight datasets demonstrate the potential of wrong-over-wrong alignment: (1) LLMs \emph{do} have preliminary capabilities to provide wrong-over-wrong judgements, with an accuracy of up to 70.9\% across the four datasets spanning knowledge, factuality, commonsense, and structured data. We further identify that score-based methods together with margin filtering might be the best approach for preference elicitation and pairs of wrong answers with the \emph{larger} gaps in scores should be retained. 
(2) Alignment with the elicited wrong-over-wrong preferences confirms the above hypotheses, demonstrating improvements across all three objectives: aligned LLMs are on average across datasets up to 9.0\% less wrong as evaluated by the proxy, up to 7.0\% of answers become correct post-alignment, while the Estimated Calibration Error (ECE) is reduced by up to 9.4\% across the four tasks. We envision wrong-over-wrong alignment as a complement to existing alignment procedures, with the unique potential to expand the frontier of model capabilities for tasks where correctness is not readily available.

\section{Methodology} \label{sec:method}

\subsection{Eliciting Wrong-over-Wrong Preferences}

% [preamble] 
When LLM-generated answers are incorrect, it is often possible that some answers are \emph{less wrong} than others. We term this ``varying shades of wrong'' and investigate whether LLMs could provide reliable wrong-over-wrong preferences if no reliable ground truths are available.

% [preliminary math] input: {LLM, q, (a1, a2)}, output: {(a1>a2) or (a1<a2)}
Formally, given a question $\boldsymbol{q}$ and a pair of \emph{wrong} answers $(\boldsymbol{a}_1, \boldsymbol{a}_2)$, we aim to employ LLMs to approximate a ``ground-truth'' wrong-over-wrong preference function $f(\boldsymbol{a}_1 \succ  \boldsymbol{a}_2 \mid \boldsymbol{q}) \rightarrow \{1, 0, -1\}$, where $1$ indicates that $\boldsymbol{a}_1$ is less wrong and should be preferred over $\boldsymbol{a}_2$, $-1$ vice versa, and $0$ indicates the two answers are not separable due to ambiguity or uncertainty. Since the ground-truth preference function $f(\boldsymbol{a}_1 \succ \boldsymbol{a}_2 \mid \boldsymbol{q})$ is usually unavailable, we propose using a silver function $\hat{f}(\boldsymbol{a}_1 \succ \boldsymbol{a}_2 \mid \boldsymbol{q})$ to estimate $f$ and employ $\hat{f}$ for evaluation  (e.g., in the shortest path problem, a path of length 8 is less wrong than a path of length 11, when 5 is correct).
% The definition is a little bit problematic...
% \begin{align*}
%     f(\boldsymbol{a}_1 \succ \boldsymbol{a}_2 \mid \boldsymbol{q}) =  
%     \begin{cases}
%         1, & \text{if } \boldsymbol{a}_1 \text{ is less wrong than } \boldsymbol{a}_2 \text{ for question } \boldsymbol{q}\\
%         0, & \text{if it is unable to determine the wrongness between } \boldsymbol{a}_1 \text{ and } \boldsymbol{a}_2\\
%         -1, & \text{if } \boldsymbol{a}_2 \text{ is less wrong than } \boldsymbol{a}_1 \text{ for question } \boldsymbol{q}
%     \end{cases}
% \end{align*}

Although using proxy functions to directly construct wrong-over-wrong preferences may seem more ideal, these proxies often rely on the availability of ground-truth answers or models trained on ground truth, which is against our setup. Moreover, not all datasets have well-defined proxies, and those that do vary vastly in format. Thus, we propose to elicit wrong-over-wrong preferences from an LLM, $f_{\textit{LLM}}(\boldsymbol{a}1 \succ \boldsymbol{a}2 \mid \boldsymbol{q}) \rightarrow \{1, 0, -1\}$. Given that LLMs are known to suffer from poor calibration in evaluation \citep{liu2024calibrating}, our first research question is: \emph{Can LLMs provide reliable wrong-over-wrong preferences}, or can any $f_{\textit{LLM}}$ be a good estimator of $\hat{f}$? We specifically explore five approaches including two that do not use LLMs  $f_{\textit{noLLM}}(\boldsymbol{a}1 \succ \boldsymbol{a}2 \mid \boldsymbol{q})$ for reference.

\paragraph{Consistency-based} For multiple sampled responses, the frequency of various answers can be interpreted as a measure of confidence and reliability \citep{wang2023selfconsistency, manakul2023selfcheckgpt, miao2024selfcheck}. Given $m$ sampled answers $\{\boldsymbol{a}_i\}_{i=1}^m$, the \underline{s}ampling \underline{r}epetition score is calculated as $\mathrm{sr}(\boldsymbol{a}_i)=\frac{1}{m}\sum_{j=1}^{m}\mathbbm{1} (\boldsymbol{a}_i = \boldsymbol{a}_j)$, representing how frequently an answer is repeated across multiple samples. Based on the consistency scores, the preference function is defined as: $f_{\textit{noLLM}}^{(\textit{c)}}(\boldsymbol{a}_1 \succ \boldsymbol{a}_2 \mid \boldsymbol{q})=\mathrm{sgn}(\mathrm{sr}(\boldsymbol{a}_1) - \mathrm{sr}(\boldsymbol{a}_2))$, assuming that the more frequent answers are more likely to be correct \citep{cole2023selectively}. Note that $f^{(\textit{c)}}$ might not be applicable to open-ended generations where there is no fixed set of candidate answers.

\paragraph{Logits-based} A well-calibrated LLM's token probabilities should reflect the reliability/trustworthiness of responses \citep{geng2024survey}, and a more probable answer is more likely to be correct. We employ Negative Log-Likelihood (NLL) to estimate the LLM’s confidence in each wrong answer, specifically $\mathrm{NLL}(\mathbf{a} \mid \mathbf{q}) = -\mathrm{log}(\prod_{t \in \mathbf{a}} p_t) = -\sum_{t \in \mathbf{a}} \mathrm{log}(p_t)$ where $t$ is a token in answer $\mathbf{a}$ and $p_t$ is its token probability. The logits-based preference function is then defined as: $f_{\textit{LLM}}^{(\textit{l})}(\boldsymbol{a}_1 \succ \boldsymbol{a}_2 \mid \boldsymbol{q})=\mathrm{sgn}(\mathrm{NLL}(\boldsymbol{a}_2 \mid \boldsymbol{q}) - \mathrm{NLL}(\boldsymbol{a}_1 \mid \boldsymbol{q}))$.

\paragraph{Pairwise comparison} As increasingly employed in LLM-as-a-judge research \citep{dubois2023alpacafarm, bai2023benchmarking}, an LLM can directly compare two answers. We prompt an LLM to reason about which answer in $(\boldsymbol{a}_1$, $\boldsymbol{a}_2)$ is less wrong (e.g., \emph{``select the preferred AI model's output for a given instruction based on their correctness''} ). The result is denoted as $\mathrm{PC}(\boldsymbol{a}_1 \succ \boldsymbol{a}_2 \mid \boldsymbol{q}, \mathrm{LLM}) \rightarrow \{1, 0, -1\}$ and the full prompt text is in Table \ref{tab:prompt1}. 

However, LLMs could be sensitive to answer order and suffer from positional biases \citep{wang2023large, zheng2023judging}. To mitigate this limitation, we employ consistency checks by flipping the order of answers, prompt the LLM to judge again, and disregard any inconsistent results.
The final pairwise comparison preference function after filtering is:
\begin{align*}
    f_{\textit{LLM}}^{(\textit{p)}}(\boldsymbol{a}_1 \succ \boldsymbol{a}_2 \mid \boldsymbol{q}) = \frac{1}{2}\big( \mathrm{PC}(\boldsymbol{a}_1 \succ \boldsymbol{a}_2 \mid \boldsymbol{q}, \mathrm{LLM}) - \mathrm{PC}(\boldsymbol{a}_2 \succ \boldsymbol{a}_1 \mid \boldsymbol{q}, \mathrm{LLM})  \big)
\end{align*}

% \begin{align*}
%     f_{\textit{LLM}}^{(p)}(\boldsymbol{a}_1 \succ \boldsymbol{a}_2 \mid \boldsymbol{q}) =
%     \begin{cases}
%         & \mathrm{PC}(\boldsymbol{a}_1 \succ \boldsymbol{a}_2 \mid \boldsymbol{q}, \mathrm{LLM}), \\
%         & \quad \text{if } \mathrm{LLM}(\boldsymbol{a}_1 \succ \boldsymbol{a}_2 \mid \boldsymbol{q}) \neq \mathrm{LLM}(\boldsymbol{a}_2 \succ \boldsymbol{a}_1 \mid \boldsymbol{q})\\
%         & 0, \quad \text{otherwise}\\
%     \end{cases}
% \end{align*}

\paragraph{Score-based} Another strategy is to let an LLM assign a score to each answer, where higher scores represent ``less wrong'' and then compare the scores \citep{bansal2024peering}. We employ the prompt \emph{``grade the correctness of each response on a 5-point scale from 0 to 5''} to obtain the LLM's score for each answer $\boldsymbol{a}$: $\mathrm{score}(\boldsymbol{a} \mid \boldsymbol{q}, \mathrm{LLM})$. The full prompt text is in Table \ref{tab:prompt2}. This scoring mechanism allows for a more fine-grained measure than pairwise comparison. We identify two important factors in this scoring-based approach:
\begin{itemize}[leftmargin=*]
    \item \textbf{Batch size}: We simultaneously include $b$ answers $\{\boldsymbol{a}_i\}_{i=1}^b$ to the same question $\boldsymbol{q}$ for the LLM to score in one prompt. Increasing $b$ allows the LLM to compare and contrast multiple answers simultaneously, potentially improving its ability to discern relative answer reliability. There is a trade-off between efficiency/calibration and context length.
    \item \textbf{Score margin}: \emph{Margin} indicates the absolute score difference between a wrong-over-wrong answer pair. Specifically, we use the $m$-percentile $M_m$ of all score margins to filter out pairs with smaller differences, as larger score margins are more trustworthy indicators of wrongness disparity \citep{li2023self, yang2024rlcd, yuan2024self} (e.g., employing $M_{\textit{50}}$ indicates that only the wrong-over-wrong pairs with the top 50\% score gaps are retained.).
\end{itemize}
To sum up, the score-based preference function is defined as:
\begin{align*}
    f_{\textit{LLM}}^{(\textit{s)}}(\boldsymbol{a}_1 \succ \boldsymbol{a}_2 \mid \boldsymbol{q}) = \mathrm{sgn}(\mathrm{score}_{\textit{diff}}) \cdot \mathbbm{1}(\mid \mathrm{score}_{\textit{diff}} \mid > M_m) \big)
\end{align*}
where $\mathrm{score}_{\textit{diff}} = \mathrm{score}(\boldsymbol{a}_1 \mid \boldsymbol{q}, \mathrm{LLM}) - \mathrm{score}(\boldsymbol{a}_2 \mid \boldsymbol{q}, \mathrm{LLM})$ and $\mathbbm{1}(\cdot)$ is the indicator function.

\paragraph{Heuristic} Research suggests that response length correlates with answer quality \citep{zhao2024long}, i.e., longer answers with detailed reasoning steps may be closer to a correct solution. We define this trivial heuristic preference function as: $f_{\textit{noLLM}}^{(\textit{h)}}(\boldsymbol{a}_1 \succ \boldsymbol{a}_2 \mid \boldsymbol{q})=\mathrm{sgn}(\mathrm{len}(\boldsymbol{a}_1) - \mathrm{len}(\boldsymbol{a}_2))$, where $\mathrm{sgn}$ is the sign function.
%\footnote{For positive numbers, $\mathrm{sgn}$ outputs 1; for negative numbers, $\mathrm{sgn}$ output -1; and for 0, $\mathrm{sgn}$ outputs 0.}

% \begin{align*}
%     f_{\textit{LLM}}^{(s)}(\boldsymbol{a}_1 \succ \boldsymbol{a}_2 \mid \boldsymbol{q}) =
%     \begin{cases}
%         &\mathrm{sgn}(\mathrm{score}_{\textit{LLM}}(\boldsymbol{a}_1 \mid \boldsymbol{q}, \text{prompt}_b) - \mathrm{score}_{\textit{LLM}}(\boldsymbol{a}_2 \mid \boldsymbol{q}, \text{prompt}_b)),\\
%         & \quad \text{if } |\mathrm{score}_{\textit{LLM}}(\boldsymbol{a}_1 \mid \boldsymbol{q}, \text{prompt}_b) - \mathrm{score}_{\textit{LLM}}(\boldsymbol{a}_2 \mid \boldsymbol{q}, \text{prompt}_b)| > Q_m\\
%         & \text{abstain}, \quad \text{otherwise}
%     \end{cases}
% \end{align*}

\subsection{Wrong-over-Wrong Alignment} \label{sec:method_alignment}
% [wrong-over-wrong dataset] Assume$D, d \in D, d -> a1 ... an, (a1,an) -> part 2.1, ...$
We hypothesize that there are valuable signals in wrong-over-wrong preferences, that LLMs could learn to distinguish answer reliability, improve model calibration, and more. In addition, the ease of producing large quantities of wrong answers makes it especially promising for generating synthetic training data. Despite being a contrived scenario, wrong-over-wrong alignment provides a controlled environment to study model behavior under the worst case when there are no correct answers at all. We also show that it is a good supplement to right-over-wrong alignment in \S \ref{sec:row}.

To this end, we first synthesize a dataset $\mathcal{D}_{\textit{WoW}}$ through the following pipeline. First, given an original dataset $\mathcal{D}_{0} = \{(\boldsymbol{q}^{\textit{(i)}}, \boldsymbol{a}_{gt}^{\textit{(i)}})\}_{i=1}^n$, we sample $m$ answers from the LLM and simulate scenarios with wrong answers only by filtering out correct ones: $\{\boldsymbol{a}_j^{\textit{(i)}}\mid \boldsymbol{a}_{j}^{\textit{(i)}} \neq \boldsymbol{a}_{gt}^{\textit{(i)}}\}\}_{j=1}^m \sim \mathrm{LLM}(\boldsymbol{q}^{\textit{(i)}})$. Then we iterate over every combination $(\boldsymbol{a}_j^{\textit{(i)}}$, $\boldsymbol{a}_k^{\textit{(i)}})$ of $\{\boldsymbol{a}_j^{\textit{(i)}}\mid \boldsymbol{a}_{j}^{\textit{(i)}} \neq \boldsymbol{a}_{gt}^{\textit{(i)}}\}\}_{j=1}^m$, and judge the answers with $f_{\textit{LLM}}$ or $f_{\textit{noLLM}}$. The one considered less wrong is the \underline{c}hosen response $\boldsymbol{a}_c$ and the other is the \underline{r}ejected response $\boldsymbol{a}_r$. We ignore the cases where the LLM gives a tie ($f_{\textit{LLM}} = 0$). Full details of wrong-over-wrong dataset construction are available in Algorithm \ref{alg:WoW_generation}.

% [alignment training] SSPO writing
With this dataset, we fine-tune an LLM using any preference optimization method \citep{azar2024general, hong2024orpo, wu2024self, calandriello2024human}. For example, the training objective of DPO can be expressed as: 
\begin{align*}
    \mathcal{L}_{\textit{DPO}}(\pi_\theta; \pi_{\textit{ref}}) = -\mathbb{E}_{(\boldsymbol{a}_c, \boldsymbol{a}_r, \boldsymbol{q}) \sim \mathcal{D}_{\textit{WoW}}} \left[ \log \sigma \left( \beta \log \frac{\pi_\theta(\boldsymbol{a}_c \mid \boldsymbol{q})}{\pi_{\textit{ref}}(\boldsymbol{a}_c \mid \boldsymbol{q})} - \log \frac{\pi_\theta(\boldsymbol{a}_r \mid \boldsymbol{q})}{\pi_{\textit{ref}}(\boldsymbol{a}_r \mid \boldsymbol{q})}\right) \right]
\end{align*}
where $\pi_\theta$ is the policy model and $\pi_{\textit{ref}}$ is the reference policy. To evaluate our hypothesis about the benefits of wrong-over-wrong alignment, we measure model performance from three aspects:

\paragraph{Less wrong} We evaluate the wrongness of answers that are still wrong post-alignment based on $\hat{f}(\boldsymbol{a}_1 \succ  \boldsymbol{a}_2 \mid \boldsymbol{q})$. A reduction in wrongness indicates that LLMs are producing a better attempt at a challenging problem. As LLMs learn to separate varying shades of wrong in the alignment tuning process, we expect the answer distribution of LLMs to move towards less-wrong direction.

\paragraph{More correct} We also evaluate the accuracy of models producing correct answers. An increase in accuracy demonstrates that surprisingly, training on wrong answers \emph{only} can guide models to produce correct answers. As the model refines its wrong-over-wrong separation, answers that were previously close-to-correct may be adjusted enough to align with the correct solution, even though the training process only employs incorrect answers.

\paragraph{Better calibrated} By training the model to prefer less-wrong answers, it learns to make finer distinctions between varying shades of wrongness and its confidence estimation should become more reliable. We evaluate the calibration of models based on token probabilities.

\section{Experiment Settings}

\paragraph{Models} We employ three open and proprietary LLMs for experiments spanning different scales and access levels. First, we use \textsc{\textsc{LLaMA3-8B}} \citep{dubey2024llama}, \textsc{GPT-3.5}, and \textsc{GPT-4o} \citep{achiam2023gpt} to sample 10 answers per problem with random option orders for multiple-choice questions. We employ a temperature of 1.0 and a max generation length of 1024. LLMs employed in this stage are called \emph{generators}. We then elicit wrong-over-wrong preferences using the same LLMs with each of $f_{\textit{noLLM}}^{(\textit{\textit{h})}}$, $f_{\textit{noLLM}}^{(\textit{c})}$, $f_{\textit{LLM}}^{(\textit{l})}$, $f_{\textit{LLM}}^{(\textit{p})}$, and $f_{\textit{LLM}}^{(\textit{s})}$. LLMs employed in this stage are called \emph{evaluators}. In Appendix \ref{sec:more_llm} we also investigate the performance of \textsc{Gemini-flash}, \textsc{Gemini-pro} \citep{team2023gemini}, \textsc{Mistral-7B} \citep{jiang2023mistral}, and \textsc{Gemma-7B} \citep{team2024gemma}.

Finally, we conduct QLoRA fine-tuning \citep{dettmers2023qlora} on \textsc{LLaMA3-8B} using the collected wrong-over-wrong preferences through DPO. We experiment on preferences elicited from 4 methods, respectively $f_{\textit{GPT-4o}}^{(\textit{p})}$ with consistency checks, $f_{\textit{GPT-4o}}^{(\textit{s})}$ with $M_{\textit{50}}$ and $M_{\textit{10}}$, and $\hat{f}$ as an oracle for reference. We sample 4000 wrong-over-wrong pairs for each experiment.

\paragraph{Data} We employ datasets with a reasonable but not necessarily \emph{perfect} proxy correctness function $p(\boldsymbol{a} \mid \boldsymbol{q})$. Specifically, the proxy function gives a heuristic-based or model-based correctness score to each answer based on ground-truth. The silver ground-truth of wrong-over-wrong preference can then be calculated as: 
\begin{align*} 
    \hat{f}(\boldsymbol{a}_1 \succ \boldsymbol{a}_2 \mid \boldsymbol{q}) = \mathrm{sgn} \big(p(\boldsymbol{a}_1 \mid \boldsymbol{q}) - p(\boldsymbol{a}_2 \mid \boldsymbol{q})\big) 
\end{align*} 
% what is this] [what is the proxy for this and why] [details (optional)]
\begin{itemize}[leftmargin=*]
    \item \textbf{Knowledge Crosswords (KC)} \citep{ding2023knowledge} is a multiple-choice structured knowledge reasoning benchmark where LLMs fill in 3 blanks of entities with structured factual constraints. The correctness proxy is $p^{(\textit{kc)}}(\boldsymbol{a} \mid \boldsymbol{q}) = \frac{\text{\# of correctly filled blanks}}{\text{\# of all blanks} (3)}$.
    \item \textbf{NLGraph (NLG)} \citep{wang2023can} is a graph reasoning benchmark and we employ the shortest path subset, where LLMs find the shortest path between two nodes in a weighted undirected graph. The correctness proxy is $p^{(\textit{sp)}}(\boldsymbol{a} \mid \boldsymbol{q}) = 1-\frac{|w_{\boldsymbol{a}} - w_{\textit{gt}}|}{w_{\textit{worst}} - w_{\textit{gt}}}$, with $w_{\boldsymbol{a}}$ being the weight of the shortest path in $\boldsymbol{a}$, $w_{\textit{gt}}$ being ground truth, and $w_{\textit{worst}}$ being the weight of the longest simple path.
    \item \textbf{Bio Generation (BG)} LLMs are asked to generate a biography of a named individual, and the correctness of the answer is evaluated via FActScore \citep{min2023factscore}, which computes the percentage of atomic facts supported by retrieved passages. The names and retrieved passages are sourced from Wikipedia. The proxy function is defined as $p^{(\textit{bg)}}(\boldsymbol{a} \mid \boldsymbol{q}) = \mathrm{FActScore}(\boldsymbol{a} \mid \boldsymbol{q}) \in [0,1]$.
    \item \textbf{COM$^2$} \citep{fang2024complex} is a multiple-choice commonsense reasoning benchmark. We use Vera \citep{liu2023vera} to obtain the commonsense plausibility scores for each option. The proxy function is defined as $p^{(\textit{cs)}}(\boldsymbol{a} \mid \boldsymbol{q}) = \mathrm{Vera}(\boldsymbol{a} \mid \boldsymbol{q}) \in [0,1]$.
\end{itemize}
We sample 625, 625, 625, and 380 questions from each dataset, each split into training sets $\mathcal{D}_{\textit{train}}$, validation sets $\mathcal{D}_{\textit{val}}$, and test sets $\mathcal{D}_{\textit{test}}$ with an approximately 8:1:1 ratio. We then generate $\mathcal{D}_{\textit{WoW}}$ from $\mathcal{D}_{\textit{train}}$ employing the pipeline in \S\ref{sec:method_alignment}. $\mathcal{D}_{\textit{test}}$ is used for evaluation and $\mathcal{D}_{\textit{val}}$ for hyperparameter tuning. We also experiment on datasets without clear right-wrong distinctions in Appendix \ref{sec:no_proxy}.

\paragraph{Evaluation and Metrics} 
For the first research question of eliciting wrong-over-wrong preferences, we compare against the proxy ground-truth preferences $\hat{f}(\boldsymbol{a}_1 \succ \boldsymbol{a}_2 \mid \boldsymbol{q})$. The accuracy of LLM-generated wrong-over-wrong preferences is calculated as:
\begin{align*}
    \mathrm{Acc}_\textit{WoW} = \frac{1}{|\mathcal{D}_{\textit{WoW}}|}\sum\nolimits_{(\boldsymbol{q},\boldsymbol{a}_c,\boldsymbol{a}_r) \in \mathcal{D}_{\textit{WoW}}} \mathbbm{1} (\hat{f}(\boldsymbol{a}_c \succ \boldsymbol{a}_r \mid \boldsymbol{q}) = f_{\textit{(no)LLM}}(\boldsymbol{a}_c \succ \boldsymbol{a}_r \mid \boldsymbol{q}))
\end{align*}
For the second research question of wrong-over-wrong alignment, we evaluate improvements across three objectives after aligning with $\mathcal{D}_{\textit{WoW}}$:
%For this, we sample 10 answers to each test question in $\mathcal{D}_{\textit{test}}=\{(\boldsymbol{q}^{(i)}, \boldsymbol{a}_{gt}^{(i)}, \boldsymbol{a}_{1,2,\dots,m}^{(i)})\ \mid i=1, 2, \dots, n_{\textit{test}}\}$. 
\begin{itemize}[leftmargin=*]
    \item \textbf{Less Wrong}: We employ the average proxy function scores of all wrong answers: $p_{\textit{wrong}} = \frac{1}{N_{\textit{wrong}}}\sum\nolimits_{ \mathcal{D}_{\textit{test}}} \sum\nolimits_{\boldsymbol{a}_i \neq \boldsymbol{a}_{\textit{gt}}} p(\boldsymbol{a}_i\mid \boldsymbol{q})$, where $N_{\textit{wrong}}$ is the number of wrong answers and $\boldsymbol{a}_{\textit{gt}}$ indicates the correct answer.
    \item \textbf{More Correct}: We employ task accuracy: $\mathrm{Acc} = \frac{1}{N}\sum\nolimits_{ \mathcal{D}_{\textit{test}}} \sum\nolimits_{\boldsymbol{a}_i} \mathbbm{1}(\boldsymbol{a}_i = \boldsymbol{a}_{gt})$, where $N$ is the number of all answers.
    \item \textbf{Better Calibration}: We extract the confidence of LLM answers with $\mathrm{exp}(\mathrm{NLL}(\mathbf{a} \mid \mathbf{q})) \in [0,1]$, create 10 bins of $[0.1i, 0.1(i+1)]$, and employ Expected Calibration Error (ECE) \citep{guo2017calibration} to measure the calibration quality.
\end{itemize}

%\paragraph{Implementation} [hyperparameters] [implementation details]

\section{Results}

\begin{table}[t]\centering
% \scriptsize
\setlength{\tabcolsep}{1.5pt}
\resizebox{0.9\textwidth}{!}{
\begin{tabular}{lccccc|cccc|cccc|c}\toprule[1.5pt]
\multirow{2}{*}{\textbf{Method}}& \multirow{2}{*}{\textbf{Margin}}&\multicolumn{4}{c}{\textsc{LLaMA3-8B}} &\multicolumn{4}{c}{\textsc{GPT-3.5}} &\multicolumn{4}{c}{\textsc{GPT-4o}} &\multirow{3}{*}{Overall} \\
% \cmidrule{3-14}
 & &KC &BG &COM$^2$ &NLG &KC &BG &COM$^2$ &NLG &KC &BG &COM$^2$ &NLG & \\
\midrule[0.75pt]\multicolumn{15}{c}{\textit{\ \ \textbf{\textsc{Evaluator-Independent}}} }\\ \midrule[0.75pt] 
\multirow{2}{*}{\textsc{Heuristic}} &$M_{\textit{50}}$ &.483 &.480 &.498 &.492 &.546 &.519 &.482 &.463 &.491 &.459 &.444 &.503 &.488 \\ 
 &$M_{\textit{10}}$ &.502 &.425 &.492 &.514 &.589 &.568 &.489 &.420 &.500 &.408 &.380 &.533 &.474 \\
\multirow{2}{*}{\textsc{Consistency}} &$M_{\textit{50}}$ &.500 & - &.505 &.582 &.526 & - &.470 &.605 &.434 & - &.565 &.548 &.559 \\
&$M_{\textit{10}}$ &.447 & - &.441 &.578 &.506 & - &.556 &\textbf{.695} &.600 & - &.423 &.494 &.566 \\
\midrule[0.75pt]\multicolumn{15}{c}{\textit{\ \ \textbf{\textsc{LLaMA3-8B as Evaluator}}} }\\ \midrule[0.75pt] 
\multirow{2}{*}{\textsc{Pairwise}} &all &.498 &.492 &.455 &.486 &.481 &.488 &.530 &.533 &.509 &.499 &.468 &.503 &.496 \\
&filter &.518 &.566 &.437 &.510 &.528 &.479 &.549 &.582 &.564 &.419 &.491 &.525 &.533 \\
\multirow{2}{*}{\textsc{Logits}}&$M_{\textit{50}}$ &.541 &.568 &.524 &.505 &.570 &.531 &.482 &.464 &.566 &.491 &.419 &.577 &.532 \\
&$M_{\textit{10}}$ &.559 &.669 &.432 &.528 &.571 &\underline{.649} &.496 &.427 &.400 &.444 &.310 &\underline{.630} &.582 \\
\multirow{2}{*}{\textsc{Score}}&$M_{\textit{50}}$ &.621 &.552 &.493 &.554 &.643 &.502 &.574 &.580 &\underline{.604} &.514 &\underline{.626} &.424 &.546 \\
&$M_{\textit{10}}$ &.654 &.551 &.458 &.579 &.701 &.485 &.659 &.524 &\textbf{.800} &\textbf{.632} &\textbf{.662} &.500 &.558 \\
\midrule[0.75pt]\multicolumn{15}{c}{\textit{\ \ \textbf{\textsc{GPT-3.5 as Evaluator}} }}\\ \midrule[0.75pt] 
\multirow{2}{*}{\textsc{Pairwise}}&all &.512 &.493 &.472 &.500 &.504 &.474 &.531 &.512 &.463 &.492 &.466 &.502 &.494 \\
&filter &.531 &.631 &.502 &.231 &.520 &.500 &.493 &.531 &.387 &.437 &.433 &.400 &.536 \\
\multirow{2}{*}{\textsc{Logits}}&$M_{\textit{50}}$ & - & - & - & - &.548 &.511 &.570 &.475 & - & - & - & - &.507 \\
&$M_{\textit{10}}$ & - & - & - & - &.541 &.538 &.570 &.430 & - & - & - & - &.505 \\
\multirow{2}{*}{\textsc{Score}} &$M_{\textit{50}}$ &.424 &.566 &.473 &.552 &.578 &.522 &.608 &.600 &.264 &.502 &.503 &.551 &.547 \\
&$M_{\textit{10}}$ &.585 &.632 &.517 &.555 &.583 &.550 &\textbf{.718} &.575 &.200 &.546 &\textbf{.662} &.573 &.590 \\
\midrule[0.75pt]\multicolumn{15}{c}{\textit{\ \ \textbf{\textsc{GPT-4o as Evaluator}}} }\\ \midrule[0.75pt] 
\multirow{2}{*}{\textsc{Pairwise}} &all &.605 &.593 &.507 &.551 &.646 &.512 &.515 &.577 &.434 &.501 &.526 &.537 &.562 \\
&filter &.691 &\underline{.689} &.533 &.602 &.712 &.536 &.558 &.661 &.417 &.490 &.604 &.549 &.624\\
\multirow{2}{*}{\textsc{Logits}}&$M_{\textit{50}}$ & - & - & - & - & - & - & - & - &.491 &.539 &.486 &.572 &.544 \\
&$M_{\textit{10}}$ & - & - & - & - & - & - & - & -&.200 &.584 &.507 &.591 &.574 \\
\multirow{2}{*}{\textsc{Score}}&$M_{\textit{50}}$ &\underline{.733} &.677 &\textbf{.544} &\underline{.605} &\underline{.793} &.591 &.617 &.661 &.547 &.520 &.581 &\textbf{.639} &\underline{.641} \\
&$M_{\textit{10}}$ &\textbf{.793} &\textbf{.795} &\underline{.534} &\textbf{.652} &\textbf{.835} &\textbf{.655} &\underline{.711} &\underline{.684} &.400 &\underline{.586} &.520 &.578 &\textbf{.709} \\
\bottomrule[1.5pt]
\end{tabular}
}
\caption[]{Accuracy of LLM-generated wrong-over-wrong preferences $Acc_\textit{WoW}$. The three LLMs across on top are employed to generate answers (\emph{generator}). Best results are in \textbf{bold}, second best are in \underline{underline}, and incompatible or unavailable results are denoted as ``-''.\footnotemark \ For pairwise comparison, margin ``all'' is without consistency checks and margin ``filter'' is with consistency checks. Overall accuracy is a weighted average across all datasets by the number of wrong-over-wrong pairs in each dataset. LLMs \emph{do} have preliminary capabilities to distinguish varying shades of wrong with up to $\mathrm{Acc}_{\textit{WoW}}$ = 0.709 achieved by the score-based approach, specifically $f_{\textit{GPT-4o}}^{(\textit{s})}$ with $M_{\textit{10}}$.}
\label{tab:res_preference}
\vspace{-5pt}
\end{table}
\footnotetext{Answers for Bio Generation questions are open-ended and we cannot compute repetition; \textsc{GPT-3.5} and \textsc{GPT-4o} are not open models for calculating logits and NLL.}

\paragraph{Eliciting Wrong-over-Wrong Preferences} We present the accuracy of LLM-generated wrong-over-wrong preferences in Table \ref{tab:res_preference}.

\begin{itemize}[leftmargin=*]
    \item \textbf{Eliciting wrong-over-wrong preference from LLMs is feasible.} Most (approach, LLM) combinations yield wrong-over-wrong preferences that are significantly better than random guess: the average $\mathrm{Acc}_\textit{WoW}$ is 0.553 across datasets, and the best $\mathrm{Acc}_\textit{WoW}$ is 0.709 achieved by $f_{\textit{GPT-4o}}^{(\textit{s})}$ with a margin of $M_{\textit{10}}$. This suggests that LLMs, with the right approach, \emph{do} possess preliminary capabilities to distinguish various shades of wrong.
    \item \textbf{Knowledge-based tasks are easier while commonsense is most challenging.} The average $\mathrm{Acc}_{\textit{WoW}}$ across all methods for Knowledge Crosswords, Bio Generation, COM$^2$, and NLGraph are 0.578, 0.560, 0.522, and 0.540 respectively. This suggests that LLMs have varying wrong-over-wrong judgement capabilities across domains: performance is best on Knowledge Crosswords, a multi-hop knowledge QA dataset, while worst on COM$^2$, a commonsense reasoning task. The ability to determine wrong-over-wrong may be impacted by LLMs' base capability on respective task; Figure \ref{fig:corr1} shows a positive correlation between model accuracy on the task and $\mathrm{Acc}_{\textit{WoW}}$.
    \item \textbf{Scored-based elicitation is the best method.} $f_{\textit{GPT-4o}}^{(\textit{s})}$ with $M_{\textit{10}}$ achieves the best $\mathrm{Acc}_{\textit{WoW}}$. The average $\mathrm{Acc}_{\textit{WoW}}$ across LLMs for consistency-based with $M_{\textit{10}}$, logits-based with $M_{\textit{10}}$, pairwise comparison with consistency checks, score-based with $M_{\textit{10}}$, and heuristic are 0.554, 0.566, 0.564, 0.619, and 0.474, indicating that the score-based method outperforms all other approaches by at least 9.4\%. This suggests that the token probabilities of aligned LLMs might not be well-calibrated \citep{sorensen2024roadmap, feng2024modular} for wrong-over-wrong contexts, while LLM-as-a-judge methodologies \citep{li2024generative} with score-based prompting offer the most promising solution.
    \item \textbf{Consistency checks and score margins are helpful.} For the pairwise comparison method, we notice that LLMs are sensitive to response order and applying consistency checks by removing inconsistent judgements due to order-flipping improves 9.0\% $\mathrm{Acc}_{\textit{WoW}}$ on average. Employing score margins to only consider the most separable cases also improves $\mathrm{Acc}_{\textit{WoW}}$ for logits-based, consistency-based, and score-based methods, with average improvements across datasets from $M_{\textit{50}}$ to $M_{\textit{10}}$ being 2.1\%, 1.3\%, and 6.9\%, and from $M_{\textit{100}}$ to $M_{\textit{10}}$ being 5.8\%, 72.4\%, and 65.3\%. The significant improvement from $M_{\textit{100}}$ to $M_{\textit{10}}$ is due to prevalent cases where the LLM gives the same score for wrong answers with different levels of wrongness.
    \item \textbf{Self-evaluation is subpar.} LLMs may not be good at producing accurate wrong-over-wrong preferences on their own generation. While \textsc{GPT-4o} provides good wrong-over-wrong preferences on answers generated by \textsc{LLaMA3-8B} and \textsc{GPT-3.5}, accuracy suffers for its own answers, with a 21.1\% and 19.7\% drop for score-based with $M_{10}$ and pairwise comparison with consistency checks on average across datasets. This echos findings that LLM self-critique and self-correction might not be satisfactory \citep{valmeekam2023can, west2023generative, huang2024large}. More interestingly, employing weak LLMs to evaluate strong LLMs may be helpful \citep{khan2024debating}, with \textsc{LLaMA3-8B} achieving 0.649 $\mathrm{Acc}_{\textit{wow}}$ on answers generated by \textsc{GPT-4o}.
\end{itemize}
\begin{table}[t]
\centering
% \scriptsize
\setlength{\tabcolsep}{1.3pt}
\resizebox{0.9\textwidth}{!}{
\begin{tabular}
{lcccccccccccc}\toprule[1.5pt]
\multirow{2}{*}{\textbf{Method}} &\multicolumn{3}{c}{KC} &\multicolumn{3}{c}{BG} &\multicolumn{3}{c}{COM$^2$} &\multicolumn{3}{c}{NLG} \\
\cmidrule(lr){2-4} \cmidrule(lr){5-7}\cmidrule(lr){8-10}\cmidrule(lr){11-13}
% \cmidrule{2-13}
&$p_{\textit{wrong}\uparrow}$ &$\mathrm{Acc}_{\uparrow}$ &$\mathrm{ECE}_{\downarrow}$ &$p_{\textit{wrong}\uparrow}$ &$\mathrm{Acc}_{\uparrow}$ &$\mathrm{ECE}_{\downarrow}$ &$p_{\textit{wrong}\uparrow}$ &$\mathrm{Acc}_{\uparrow}$ &$\mathrm{ECE}_{\downarrow}$ &$p_{\textit{wrong}\uparrow}$ &$\mathrm{Acc}_{\uparrow}$ &$\mathrm{ECE}_{\downarrow}$ \\\midrule[0.75pt]
\textsc{Original} &.466 &.555 &.235 &.532 &.027 &.576 &.312 &.669 &.053 &.750 &.142 &.649 \\ \midrule[0.75pt]
\multicolumn{13}{c}{\textit{\ \ \textbf{\textsc{Self-Generator}}} }\\ \midrule[0.75pt]
\textsc{Pairwise} filter &\cellcolor[HTML]{d4edda}.475 &\cellcolor[HTML]{d4edda}\textbf{.627} &\cellcolor[HTML]{d4edda}\textbf{.096} &\cellcolor[HTML]{d4edda}.670 &\cellcolor[HTML]{d4edda}.059 &\cellcolor[HTML]{d4edda}.500 &\cellcolor[HTML]{d4edda}.326 &\cellcolor[HTML]{d4edda}.690 &\cellcolor[HTML]{d4edda}.049 &\cellcolor[HTML]{d4edda}.806 &\cellcolor[HTML]{d4edda}.179 &\cellcolor[HTML]{d4edda}\underline{.493} \\
\textsc{Score} $M_{\textit{50}}$ &\cellcolor[HTML]{d4edda}.529 &\cellcolor[HTML]{d4edda}\underline{.597} &.251 &\cellcolor[HTML]{d4edda}.661 &\cellcolor[HTML]{d4edda}.043 &.580 &\cellcolor[HTML]{d4edda}.325 &.660 &\cellcolor[HTML]{d4edda}\underline{.039} &\cellcolor[HTML]{d4edda}.800 &\cellcolor[HTML]{d4edda}.203 &\cellcolor[HTML]{d4edda}.551 \\
\textsc{Score} $M_{\textit{10}}$ &\cellcolor[HTML]{d4edda}.532 &\cellcolor[HTML]{d4edda}.584 &.315 &\cellcolor[HTML]{d4edda}.682 &\cellcolor[HTML]{d4edda}.075 &\cellcolor[HTML]{d4edda}.561 &\cellcolor[HTML]{d4edda}\textbf{.357} &\cellcolor[HTML]{d4edda}.681 &\cellcolor[HTML]{d4edda}\textbf{.020} &\cellcolor[HTML]{d4edda}\textbf{.847} &\cellcolor[HTML]{d4edda}\underline{.292} &\cellcolor[HTML]{d4edda}.578 \\
\textsc{oracle} &\cellcolor[HTML]{d4edda}.529 &\cellcolor[HTML]{d4edda}.576 &.279 &\cellcolor[HTML]{d4edda}\textbf{.695} &\cellcolor[HTML]{d4edda}\underline{.108} &\cellcolor[HTML]{d4edda}\textbf{.440} &\cellcolor[HTML]{d4edda}.330 &\cellcolor[HTML]{d4edda}.689 &.064 &\cellcolor[HTML]{d4edda}\underline{.846} &\cellcolor[HTML]{d4edda}.182 &\cellcolor[HTML]{d4edda}.596 \\\midrule[0.75pt]
\multicolumn{13}{c}{\textit{\ \ \textbf{\textsc{Mix-Generator}}} }\\ \midrule[0.75pt]

\textsc{Pairwise} filter &\cellcolor[HTML]{d4edda}\underline{.533} &\cellcolor[HTML]{d4edda}.574 &\cellcolor[HTML]{d4edda}.201 &\cellcolor[HTML]{d4edda}.634 &\cellcolor[HTML]{d4edda}.075 &\cellcolor[HTML]{d4edda}.535 &\cellcolor[HTML]{d4edda}\underline{.355} &\cellcolor[HTML]{d4edda}\textbf{.698} &\cellcolor[HTML]{d4edda}.048 &\cellcolor[HTML]{d4edda}.832 &\cellcolor[HTML]{d4edda}.192 &\cellcolor[HTML]{d4edda}.538 \\
\textsc{Score} $M_{\textit{50}}$ &\cellcolor[HTML]{d4edda}.528 &\cellcolor[HTML]{d4edda}.590 &\cellcolor[HTML]{d4edda}\underline{.175} &\cellcolor[HTML]{d4edda}.619 &\cellcolor[HTML]{d4edda}.065 &\cellcolor[HTML]{d4edda}.523 &\cellcolor[HTML]{d4edda}.329 &.669 &.067 &\cellcolor[HTML]{d4edda}.827 &\cellcolor[HTML]{d4edda}.221 &\cellcolor[HTML]{d4edda}.585 \\
\textsc{Score} $M_{\textit{10}}$ &\cellcolor[HTML]{d4edda}.520 &\cellcolor[HTML]{d4edda}.565 &.273 &\cellcolor[HTML]{d4edda}.687 &\cellcolor[HTML]{d4edda}\textbf{.129} &\cellcolor[HTML]{d4edda}.560 &\cellcolor[HTML]{d4edda}.346 &\cellcolor[HTML]{d4edda}.677 &.065 &\cellcolor[HTML]{d4edda}.843 &\cellcolor[HTML]{d4edda}\textbf{.303} &\cellcolor[HTML]{d4edda}.522 \\
\textsc{oracle} &\cellcolor[HTML]{d4edda}\textbf{.537} &\cellcolor[HTML]{d4edda}.581 &\cellcolor[HTML]{d4edda}.185 &\cellcolor[HTML]{d4edda}\underline{.691} &\cellcolor[HTML]{d4edda}.086 &\cellcolor[HTML]{d4edda}\underline{.472} &\cellcolor[HTML]{d4edda}.328 &\cellcolor[HTML]{d4edda}\underline{.697} &.067 &\cellcolor[HTML]{d4edda}.832 &\cellcolor[HTML]{d4edda}.226 &\cellcolor[HTML]{d4edda}\textbf{.474} \\
\bottomrule[1.5pt]
\end{tabular}
}
\caption{Evaluation of wrong-over-wrong alignment on less wrong ($p_{\textit{wrong}}$), more correct ($\mathrm{Acc}$), and better calibration ($\mathrm{ECE}$). The best results are in \textbf{bold}, second best are in \underline{underline}, and \colorbox[HTML]{d4edda}{green background} indicates improvement over the original \textsc{LLaMA3-8B}. ``Self-Generator'' indicates that wrong-over-wrong pairs are generated from only \textsc{LLaMA3-8B} while ``Mix-Generator'' uses all 3 LLMs' answers. ``Oracle'' means aligning with proxy “ground-truth” wrong-over-wrong preference $\hat{f}$. Wrong-over-wrong alignment is helpful across the board, with up to 0.163, 0.161, and 0.175 improvement in reducing wrongness, increasing correct answers, and improving calibration.} 
\label{tab:res_alignment}
\vspace{-5pt}
\end{table}

\paragraph{Alignment with Wrong-over-Wrong Preferences} We present the evaluation of wrong-over-wrong alignment in Table \ref{tab:res_alignment}.

\begin{itemize}[leftmargin=*]
    \item \textbf{Wrong-over-wrong alignment helps to reduce wrongness, produce correct answers, and improve calibration.} After wrong-over-wrong alignment, \textsc{LLaMA3-8B} improves on average across datasets on $\Delta p_{\textit{wrong}}$, $\Delta \mathrm{Acc}$, and $-\Delta \mathrm{ECE}$ by 0.074, 0.045, and 0.044. This validates that wrong-over-wrong alignment is moderately helpful across the three objectives. Even though we only align on wrong answers, we still magically end up making LLM generate 4.5\% more correct answers on average. This finding highlights the potential of alignment with wrong answers generated by LLMs that come in large quantities.
    \item \textbf{Effect of wrong-over-wrong alignment varies across datasets.} We observe substantial improvement on Bio Generation and NLGraph with up to 0.163, 0.161, and 0.175 for $\Delta p_{\textit{wrong}}$, $\Delta \mathrm{Acc}$, and $-\Delta\mathrm{ECE}$. However, relatively less improvement is observed on Knowledge Crosswords and COM$^2$, with up to 0.071, 0.073, and 0.138 for $\Delta p_{\textit{wrong}}$, $\Delta \mathrm{Acc}$, and $-\Delta\mathrm{ECE}$. This may be because Bio Generation and NLGraph are \emph{open-ended} questions and there is more space for improvement compared to \emph{multiple-choice} questions in Knowledge Crosswords and COM$^2$ where wrong options are limited and LLMs are confined to pre-defined options.
    \item \textbf{Score-based methods achieve the least wrong and most correct answers while pairwise comparison achieves the best calibration.} Score-based preferences with $M_{\textit{10}}$ on self-generated data achieve the best average improvement of $\Delta p_{\textit{wrong}}$ = 0.090 across datasets. $f_{\textit{GPT-4o}}^{(\textit{s})}$ with $M_{\textit{10}}$ on mix-generated data has the best average improvement of $\Delta \mathrm{Acc}$ = 0.070. This suggests that the score-based method has the best $\mathrm{Acc}_\textit{WoW}$ and is most helpful in reducing wrongness and producing correct answers. However, $f_{\textit{GPT-4o}}^{(\textit{p})}$ with consistency checks has the best average calibration improvement of $-\Delta\mathrm{ECE}$ = 0.094. We hypothesize that the margin used in score-based methods, while improving $\mathrm{Acc}_{\textit{WoW}}$ by avoiding cases with similar degrees of wrongness, also indicates the models' tendency to avoid challenging distinctions, which in turn harms calibration.
\end{itemize}

\section{Analysis}
 % why this; what do we do; results pending (repeat this for every analysis)
 
\begin{wrapfigure}{r}{0.4\linewidth}
    \centering
    % \vspace{-2in}
    \vspace{-32pt}
    \includegraphics[width=0.98\linewidth]{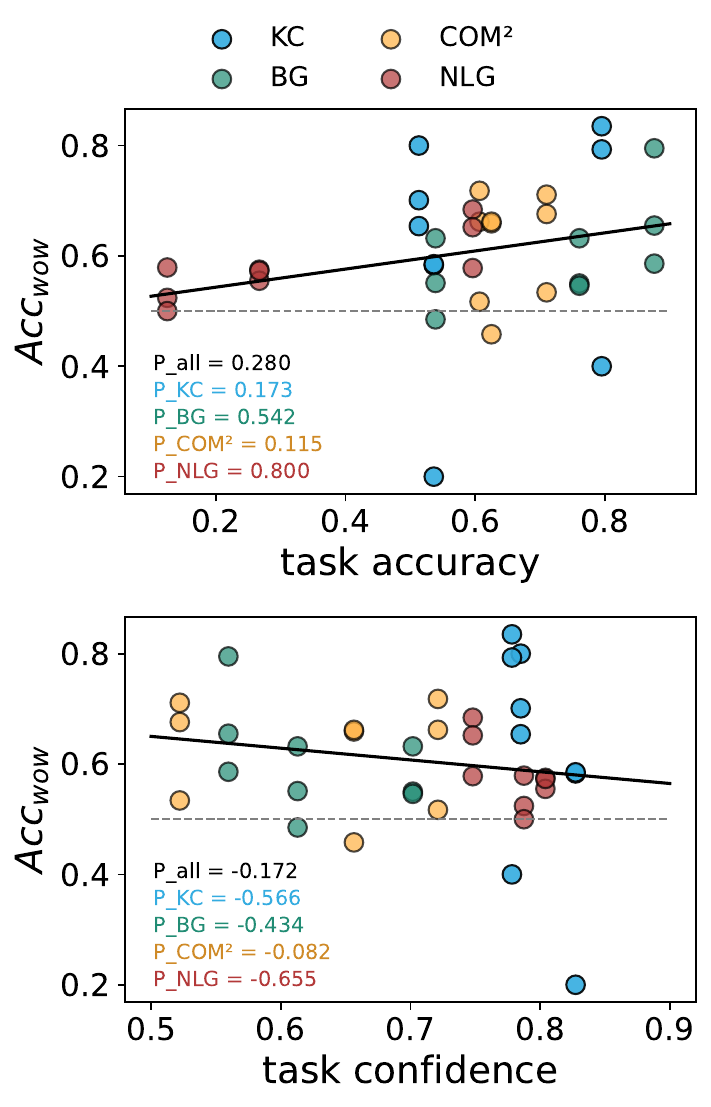}
    \vspace{-0.15in}
    \caption{Correlation between task accuracy, confidence and $\mathrm{Acc}_{\textit{WoW}}$ of score-based eliciting with $M_{\textit{10}}$. Data points are from all 3 LLMs we used to elicit wrong-over-wrong preferences. $P$ stands for Pearson correlation coefficient. The ability to elicit wrong-over-wrong preferences is positively correlated with task ability but negatively correlated with confidence.}
    \vspace{-10pt}
    \label{fig:corr1}
    \vspace{-0.3in}
\end{wrapfigure}

\paragraph{Task Utility and Wrong-over-Wrong Judgement} We explore the impact of two factors on the reliability of wrong-over-wrong judgements: accuracy of the evaluator's generated answers (\emph{task accuracy}) and the evaluator's confidence in its generated answers (\emph{task confidence}). We visualize the correlations and present Pearson correlation coefficients \citep{sedgwick2012pearson} in Figure \ref{fig:corr1}.  We observe a positive correlation between task accuracy and $\mathrm{Acc}_{\textit{WoW}}$, but a negative correlation between task confidence and $\mathrm{Acc}_{\textit{WoW}}$. This suggests that models that perform well on the task are also good at distinguishing various shades of wrong, while over-confident and under-calibrated models harm wrong-over-wrong preferences. We also find that LLM struggles to differentiate two answers with close wrongness levels in Appendix \ref{sec:wrong_margin}.

\begin{figure}[t]
    \centering
    \includegraphics[width=1\linewidth]{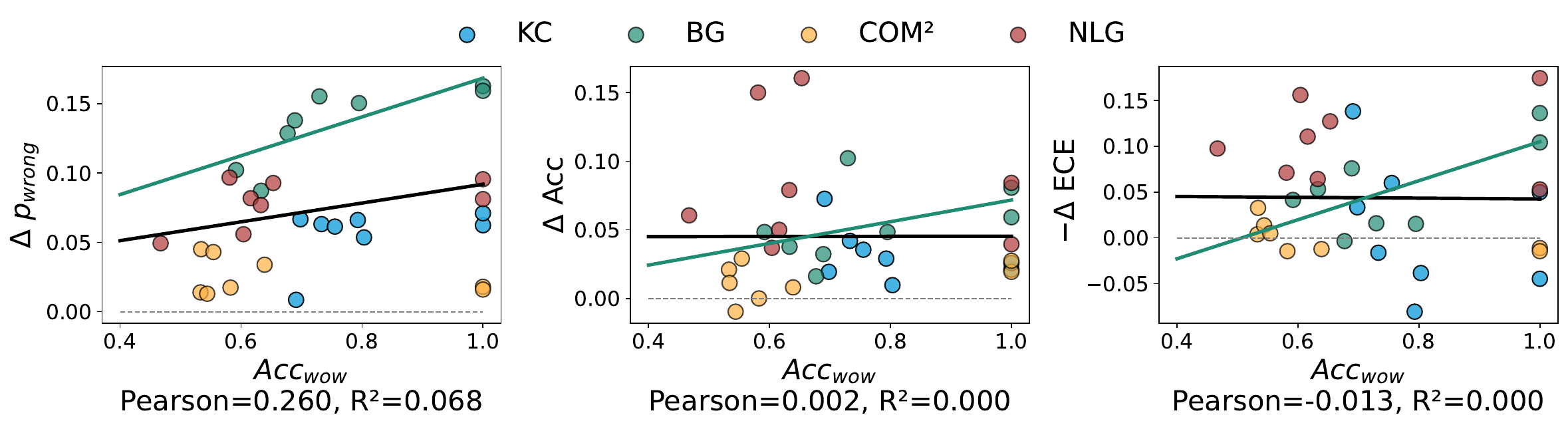}
    \vspace{-0.1in}
    \caption{Correlation between $\mathrm{Acc}_{\textit{WoW}}$ and improvement after wrong-over-wrong alignment in less wrong $\Delta p_{\textit{wrong}}$, more correct $\Delta \mathrm{Acc}$, and better calibration $-\Delta \mathrm{ECE}$. Data points are sourced from all 4 methods ($f_{\textit{GPT-4o}}^{(\textit{p})}$ with consistency checks, $f_{\textit{GPT-4o}}^{(\textit{s})}$ with $M_{\textit{50}}$ and $M_{\textit{10}}$, and oracle $\hat{f}$), and the oracle method is considered as $\mathrm{Acc}_{\textit{WoW}}$ = 1.0. The black line is the linear regression on all four datasets while the green line is the linear regression on Bio Generation dataset. Wrong-over-wrong alignment is not sensitive to the accuracy of wrong-over-wrong preference.}
    \label{fig:corr2}
    \vspace{-5pt}
\end{figure}

\paragraph{Preference Accuracy and Alignment Improvement} We examine how the quality of wrong-over-wrong judgements $\mathrm{Acc}_{\textit{WoW}}$ is related to improvements through wrong-over-wrong alignment. Figure \ref{fig:corr2} demonstrates a weak positive relationship between $\mathrm{Acc}_{\textit{WoW}}$ and improvement in $\Delta p_{\textit{wrong}}$, and no significant correlation between $\mathrm{Acc}_{\textit{WoW}}$ and improvement in $\Delta \mathrm{Acc}$ or $-\Delta \mathrm{ECE}$. Surprisingly, this suggests that improvement resulting from wrong-over-wrong alignment is nuanced and is not sensitive to the absolute accuracy of wrong-over-wrong preference data. There is also a clear positive relationship between  $\mathrm{Acc}_{\textit{WoW}}$ and $\Delta p_{\textit{wrong}}$, $\Delta \mathrm{Acc}$, and $\mathrm{ECE}$ on the Bio Generation dataset, indicating that sensitivity to wrong-over-wrong judgement quality varies by dataset.

\paragraph{Right-over-Wrong Alignment} \label{sec:row}We conduct experiments on right-over-wrong preferences and a 50:50 mix of right-over-wrong and wrong-over-wrong preferences. The results in Table \ref{tab:ana_row} reveal that: (1) Right-over-wrong alignment has the best average improvement in more correct $\Delta p_{\textit{wrong}}$ = 0.090 on mix-generated data, while on self-generated data, wrong-over-wrong alignment yields the best $\Delta p_{\textit{wrong}}$ = 0.068. Mixing generators is especially helpful to right-over-wrong alignment, with average $\Delta \mathrm{Acc}$ improved by 55.7\%. (2) Right-over-wrong alignment has the best average improvement in more correct $\Delta \mathrm{Acc}$ = 0.134, while a mixture of right-over-wrong and wrong-over-wrong preferences achieves the best average improvement in calibration $-\Delta \mathrm{ECE}$ = 0.107. This partly resolves the previous finding that alignment hurts calibration \citep{sorensen2024roadmap, feng2024modular}: we hypothesize that this is due to missing wrong-over-wrong preferences in alignment data that make LLMs unable to disentangle varying shades of wrong for fine-grained calibration.

\begin{wraptable}{r}{0.5\linewidth}
    \centering
    % \scriptsize
    \setlength{\tabcolsep}{1pt}
    \resizebox{1\linewidth}{!}{
    \begin{tabular}{lcccccccccc}\toprule[1.5pt]
    \multirow{3}{*}{\textbf{Method}} &\multicolumn{3}{c}{\textbf{KC}} &\multicolumn{3}{c}{\textbf{COM$^2$}} &\multicolumn{3}{c}{\textbf{NLG}} \\
    \cmidrule(lr){2-4}\cmidrule(lr){5-7}\cmidrule(lr){8-10}
    &$p_{\textit{wrong}\uparrow}$ &$\mathrm{Acc}_{\uparrow}$ &$\mathrm{ECE}_{\downarrow}$ &$p_{\textit{wrong}\uparrow}$ &$\mathrm{Acc}_{\uparrow}$ &$\mathrm{ECE}_{\downarrow}$ &$p_{\textit{wrong}\uparrow}$ &$\mathrm{Acc}_{\uparrow}$ &$\mathrm{ECE}_{\downarrow}$ \\\midrule[.75pt]
    \textsc{Original} &.466 &.555 &.235 &.312 &.669 &.053 &.750 &.142 &.649 \\
    \textsc{Best W}&\cellcolor[HTML]{d4edda}.529 &\cellcolor[HTML]{d4edda}.597 &.251 &\cellcolor[HTML]{d4edda}.355 &\cellcolor[HTML]{d4edda}.698 &\cellcolor[HTML]{d4edda}.048 &\cellcolor[HTML]{d4edda}.847 &\cellcolor[HTML]{d4edda}.292 &\cellcolor[HTML]{d4edda}.578 \\
    \midrule[.75pt]\multicolumn{10}{c}{\textit{\ \ \textbf{\textsc{Self-Generator}}} }\\ \midrule[.75pt]
    \textsc{r+w (pairwise)} &\cellcolor[HTML]{d4edda}.493 &\cellcolor[HTML]{d4edda}.706 &\cellcolor[HTML]{d4edda}.065 &\cellcolor[HTML]{d4edda}.359 &\cellcolor[HTML]{d4edda}.705 &.057 &\cellcolor[HTML]{d4edda}.806 &\cellcolor[HTML]{d4edda}.179 &\cellcolor[HTML]{d4edda}.493 \\
    \textsc{r+w (score $M_{\textit{50}}$)} &\cellcolor[HTML]{d4edda}.540 &\cellcolor[HTML]{d4edda}.690 &\cellcolor[HTML]{d4edda}.055 &\cellcolor[HTML]{d4edda}.335 &\cellcolor[HTML]{d4edda}.673 &\cellcolor[HTML]{d4edda}.037 &\cellcolor[HTML]{d4edda}.814 &.139 &.667 \\
    \textsc{r+w (score $M_{\textit{10}}$)} &\cellcolor[HTML]{d4edda}.503 &\cellcolor[HTML]{d4edda}.665 &\cellcolor[HTML]{d4edda}.174 &\cellcolor[HTML]{d4edda}.340 &\cellcolor[HTML]{d4edda}.687 &.070 &\cellcolor[HTML]{d4edda}.815 &.137 &.650 \\
    \textsc{r+w (oracle)} &\cellcolor[HTML]{d4edda}\textbf{.607} &\cellcolor[HTML]{d4edda}.785 &\cellcolor[HTML]{d4edda}.060 &\cellcolor[HTML]{d4edda}.357 &\cellcolor[HTML]{d4edda}.703 &.055 &\cellcolor[HTML]{d4edda}.760 &\cellcolor[HTML]{d4edda}.158 &\cellcolor[HTML]{d4edda}.549 \\
    \textsc{r} &\cellcolor[HTML]{d4edda}.579 &\cellcolor[HTML]{d4edda}\textbf{.805} &\cellcolor[HTML]{d4edda}.079 &\cellcolor[HTML]{d4edda}.373 &\cellcolor[HTML]{d4edda}.692 &\cellcolor[HTML]{d4edda}\textbf{.029} &\cellcolor[HTML]{d4edda}.777 &\cellcolor[HTML]{d4edda}.153 &.652\\
    \midrule[.75pt]\multicolumn{10}{c}{\textit{\ \ \textbf{\textsc{Mix-Generator}}} }\\ \midrule[.75pt]
    \textsc{r+w (pairwise)} &\cellcolor[HTML]{d4edda}.530 &\cellcolor[HTML]{d4edda}.705 &\cellcolor[HTML]{d4edda}.053 &\cellcolor[HTML]{d4edda}.327 &.658 &.091 &\cellcolor[HTML]{d4edda}.825 &\cellcolor[HTML]{d4edda}.218 &\cellcolor[HTML]{d4edda}.493 \\
    \textsc{r+w (score $M_{\textit{50}}$)} &\cellcolor[HTML]{d4edda}.536 &\cellcolor[HTML]{d4edda}.705 &\cellcolor[HTML]{d4edda}.094 &\cellcolor[HTML]{d4edda}.326 &\cellcolor[HTML]{d4edda}.716 &.070 &\cellcolor[HTML]{d4edda}.836 &\cellcolor[HTML]{d4edda}.263 &\cellcolor[HTML]{d4edda}.548 \\
    \textsc{r+w (score $M_{\textit{10}}$)} &\cellcolor[HTML]{d4edda}.559 &\cellcolor[HTML]{d4edda}.711 &\cellcolor[HTML]{d4edda}.104 &\cellcolor[HTML]{d4edda}.343 &.655 &\cellcolor[HTML]{d4edda}.034 &\cellcolor[HTML]{d4edda}.842 &\cellcolor[HTML]{d4edda}\textbf{.308} &\cellcolor[HTML]{d4edda}.496 \\
    \textsc{r+w (oracle)} &\cellcolor[HTML]{d4edda}.567 &\cellcolor[HTML]{d4edda}.740 &\cellcolor[HTML]{d4edda}\textbf{.025} &\cellcolor[HTML]{d4edda}.344 &\cellcolor[HTML]{d4edda}\textbf{.748} &.113 &\cellcolor[HTML]{d4edda}.826 &\cellcolor[HTML]{d4edda}.274 &\cellcolor[HTML]{d4edda}\textbf{.490} \\
    \textsc{r} &\cellcolor[HTML]{d4edda}.568 &\cellcolor[HTML]{d4edda}.787 &\cellcolor[HTML]{d4edda}.110 &\cellcolor[HTML]{d4edda}\textbf{.374} &\cellcolor[HTML]{d4edda}.711 &.183 &\cellcolor[HTML]{d4edda}\textbf{.856} &\cellcolor[HTML]{d4edda}.268 &\cellcolor[HTML]{d4edda}.537 \\
    
    \bottomrule[1.5pt]
    \end{tabular}
    }
    \caption{
    Evaluation of right-over-wrong alignment on less wrong ($p_{\textit{wrong}}$), more correct ($\mathrm{Acc}$), and better calibration ($\mathrm{ECE}$). ``\textsc{w}'' and ``\textsc{r}'' stand for wrong-over-wrong and right-over-wrong data. Wrong-over-wrong alignment is a good supplement to existing right-over-wrong alignment and best employed in conjunction.
    }
    \label{tab:ana_row}
    \vspace{-30pt}
\end{wraptable}

\paragraph{Generalization to Unseen Data} We examine how wrong-over-wrong alignment can generalize to other unseen tasks in the same domain. We employ Hellaswag \citep{zellers2019hellaswag} and the maximum flow subtask in NLGraph \citep{wang2023can} as two unseen datasets while being in the same domain as the datasets employed for alignment tuning, COM$^2$ and shortest path. Results in Table \ref{tab:ana_indomain} demonstrate that wrong-over-wrong alignment could indeed generalize to unseen data, with an average improvement of 0.118, 0.002, and 0.029 on $\Delta p_{\textit{wrong}}$, $\Delta \mathrm{Acc}$, and $-\Delta \mathrm{ECE}$.

\paragraph{Qualitative Examples} We present qualitative examples in Table \ref{tab:qualitative}, showing that WoW-aligned models could generate less wrong, more correct, and better calibrated answers. More examples are in Appendix \ref{sec:example}.

\section{Related Work}
% alignment is important ... [RLHF, PPO, reward models] -> [DPO, variant] -> [RLAIF, self-rewarding, self-play, ...] [however, WoW, we think]

\paragraph{LLM Alignment} In recent years, aligning LLMs with human preferences has become an important research question, driven by the need for safety \citep{dai2023safe, qi2024fine, huang2024safedreamer}, honesty \citep{yang2023alignment, wen2024art}, factuality\citep{liang2024learning, lin2024flame}, diversity \citep{ding2024quality}, etc. The initial methods, such as Reinforcement Learning from Human Feedback (RLHF) \citep{ziegler2019fine, stiennon2020learning, ouyang2022training, munos2023nash, chakraborty2024parl}, leverage human preferences to train reward models and employ algorithms such as Proximal Policy Optimization (PPO) \citep{schulman2017proximal} for alignment.
%These methods rely heavily on reward models distilled from human feedback to refine LLM behavior.
Recent alignment methods move away from explicit reward models, as seen in Direct Preference Optimization (DPO) \citep{rafailov2023direct}. Alongside this, Reinforcement Learning from AI Feedback (RLAIF) \citep{bai2022constitutional, lee2023rlaif} and LLM-as-a-Judge \citep{dubois2023alpacafarm, bai2023benchmarking} frameworks have introduced AI-driven feedback systems, allowing LLMs to give feedback and improve themselves with minimal human intervention. Techniques such as Self-Alignment \citep{li2023self, wang2024enabling, yuan2024self, sun2024salmon}, which involves models evaluating and aligning with their own outputs, represent a further step towards scalable oversight. Instead of alignment objective, we investigate an orthogonal research question to the improvement in alignment approaches and focus on the preference data: in challenging tasks and contexts when correct and high-quality answers are not readily available \citep{baheti2024leftover, agarwal2024onpolicy}, could LLMs distinguish various shades of wrong and align with wrong-over-wrong preferences? Prior work has also established the utility and representativeness of fine-grained and continuous preferences over categorical labels in uncertain \citep{lin2022teaching, tian2023just} and subjective contexts \citep{sorensen2024roadmap, feng2024modular}. 
% The overall experiments demonstrate that even when ``ground-truths'' are available, wrong-over-wrong alignment is still a good supplement to right-over-wrong alignment.

\begin{table}[t]
\small
\centering
\resizebox{0.9\linewidth}{!}{
\begin{tabularx}{\textwidth}{lX}
% {\textwidth}{l}
\toprule[1.5pt]
\textbf{Question} & In an undirected graph, the nodes are numbered from 0 to 4, and the edges are represented by (start node, end node, weight): (0, 2, 2), (0, 5, 4), (0, 3, 1), (0, 1, 3), (1, 2, 4), (1, 5, 1), (1, 4, 4), (2, 4, 3), (2, 3, 1), (3, 5, 2), (3, 4, 3), (4, 5, 2). Q: Give the shortest path from node 0 to node 1. Please also give the total weight of the shortest path. \textcolor{red}{[Correct answer: 4 (0 $\rightarrow$ 3 $\rightarrow$ 4).]}\\ \midrule
\textbf{Original Answer} & [CoT steps]. The final answer is:  shortest path: 0 $\rightarrow$ 2 $\rightarrow$ 3 $\rightarrow$ 4; total weight: 6.  \\ \midrule
\textbf{Less Wrong} & [CoT steps]. Therefore, the answer is: The shortest path from node 0 to node 4 is 0 $\rightarrow$ 3 $\rightarrow$ 2 $\rightarrow$ 4, with a total weight of 5.  \textcolor{red}{[The path found is shorter: 6 $\rightarrow$ 5.]} \\ \midrule
\textbf{More Correct} & [CoT steps]. Final answer: The shortest path from node 0 to node 4 is Node 0 $\rightarrow$ Node 3 $\rightarrow$ Node 4, with a total weight of 4. \textcolor{red}{[Successfully find the shortest path.]} \\ \midrule
\textbf{Better Calibrated} & [CoT steps]. The total weight of the shortest path is: 0 $\rightarrow$ 2 (weight 2), 2 $\rightarrow$ 3 (weight 1), 3 $\rightarrow$ 4 (weight 3). Total weight: 2 + 1 + 3 = 6. \textcolor{red}{[Even though the answer is equally wrong compared to the original answer, NLL increases from 0.255 $\rightarrow$ 0.373.]} \\
\bottomrule[1.5pt]
\end{tabularx}
}
\caption{Qualitative examples on NLGraph where WoW-aligned models become less wrong, more correct, and better calibrated on NLGraph. We provide explanations in \textcolor{red}{[red]}.}
\label{tab:qualitative}
\vspace{-10pt}
\end{table}

\begin{wraptable}{r}{0.5\linewidth}{
\centering
\vspace{-15pt}
% \scriptsize
% \setlength{\tabcolsep}{1.2pt}
% \renewcommand{\arraystretch}{0.8}
\resizebox{1\linewidth}{!}{
\begin{tabular}{lcccccc}\toprule[1.5pt]
\multirow{3}{*}{\textbf{Method}} &\multicolumn{3}{c}{\textbf{HellaSwag}} &\multicolumn{3}{c}{\textbf{Maximum Flow}} \\\cmidrule(lr){2-4}\cmidrule(lr){5-7}
&$p_{\textit{wrong}\uparrow}$ &$\mathrm{Acc}_{\uparrow}$ &$\mathrm{ECE}_{\downarrow}$ &$p_{\textit{wrong}\uparrow}$ &$\mathrm{Acc}_{\uparrow}$ &$\mathrm{ECE}_{\downarrow}$  \\
\midrule[0.75pt]
\textsc{Original} &.230 &.737 &.089 &.112 &.069 &.663 \\
\midrule[.75pt]
\multicolumn{7}{c}{\textit{\ \ \textbf{\textsc{Self-Generator}}} }\\ \midrule[.75pt]
\textsc{pairwise} &\cellcolor[HTML]{d4edda}.243 &.729 &.098 &\cellcolor[HTML]{d4edda}.344 &.066 &\cellcolor[HTML]{d4edda}\textbf{.567} \\
\textsc{score} $M_{\textit{50}}$ &.220 &.679 &\cellcolor[HTML]{d4edda}.047 &\cellcolor[HTML]{d4edda}.330 &\cellcolor[HTML]{d4edda}.083 &\cellcolor[HTML]{d4edda}.637 \\
\textsc{score} $M_{\textit{10}}$ &\cellcolor[HTML]{d4edda}.264 &.719 &\cellcolor[HTML]{d4edda}.068 &\cellcolor[HTML]{d4edda}.342 &\cellcolor[HTML]{d4edda}\textbf{.109} &.673 \\
\textsc{oracle} &.227 &.729 &\cellcolor[HTML]{d4edda}{.023} &\cellcolor[HTML]{d4edda}.151 &.049 &\cellcolor[HTML]{d4edda}.659 \\
\midrule[.75pt] \multicolumn{7}{c}{\textit{\ \ \textbf{\textsc{Mix-Generator}}} }\\ \midrule[.75pt]
\textsc{pairwise} &\cellcolor[HTML]{d4edda}.247 &.706 &\cellcolor[HTML]{d4edda}.082 &\cellcolor[HTML]{d4edda}\textbf{.348} &\cellcolor[HTML]{d4edda}.094 &\cellcolor[HTML]{d4edda}.533 \\
\textsc{score} $M_{\textit{50}}$ &.204 &.729 &.090 &\cellcolor[HTML]{d4edda}.346 &\cellcolor[HTML]{d4edda}.083 &\cellcolor[HTML]{d4edda}.621 \\
\textsc{score} $M_{\textit{10}}$ &\cellcolor[HTML]{d4edda}.250 &\cellcolor[HTML]{d4edda}\textbf{.771} &.117 &\cellcolor[HTML]{d4edda}.326 &\cellcolor[HTML]{d4edda}.089 &\cellcolor[HTML]{d4edda}.627 \\
\textsc{oracle} &\cellcolor[HTML]{d4edda}\textbf{.267} &\cellcolor[HTML]{d4edda}.753 &.099 &\cellcolor[HTML]{d4edda}.202 &\cellcolor[HTML]{d4edda}.074 &\cellcolor[HTML]{d4edda}.582 \\
\bottomrule[1.5pt]
\end{tabular}}
\vspace{-0.1in}
\caption{Generalization to unseen datasets, from COM$^2$ to HellaSwag and from shortest path to maximum flow. While tuned only on COM$^2$ and shortest path, the aligned models improve on the two unseen datasets as well, with an average improvement of 0.118, 0.002, and 0.029 on $\Delta p_{\textit{wrong}}$, $\Delta \mathrm{Acc}$, and $-\Delta \mathrm{ECE}$.}
\label{tab:ana_indomain}
}
\vspace{-10pt}
\end{wraptable}

\paragraph{Easy-to-Hard Generalization} Most of these advancements in alignment benefit from the availability of ``correct'' answers. In contrast, our work posits that ``correct'' answers are inevitably unavailable or unreliable for emerging and challenging tasks. \citet{bowman2022measuring} points out advanced models may engage in tasks that are beyond current evaluation capabilities. Easy-to-hard generalization \citep{burns2023weak, hase2024unreasonable, xu2024towards, sun2024easy} introduces a series of solutions hoping to improve LLMs on hard tasks by finetuning on easy in-domain tasks. Our work uniquely points out a new path for training future LLMs on hard tasks when correct answers are unavailable, arguing for wrong-over-wrong alignment based on synthetic preference data as a complementary training objective.

\section{Conclusion}
With the growing race towards bigger, better LLMs capable of solving a wider range of tasks, it becomes evident that the availability of carefully curated data is a major bottleneck. Our work investigates the potential to alleviate this limitation by eliciting preferences among wrong answers with an LLM and aligning with these wrong-over-wrong preferences. We empirically investigate wrong-over-wrong alignment with seven LLMs and eight datasets. We find that LLMs \emph{do} have preliminary capability to rank wrong answers and produce reliable wrong-over-wrong preferences. The strongest approach such as score-based LLM-as-a-judge achieves up to 70.9\% accuracy across datasets. In addition, alignment with such wrong-over-wrong preferences is helpful in calibrating the models, e.g., reducing answer ``wrongness'' by 9.0\%, producing 7.0\% more correct answers, and improving overall model calibration by 9.4\% on average. We envision future wrong-over-wrong alignment methods as an important addition to advance LLM capabilities on challenging tasks where correct answers or rich annotations are not readily available.

\section*{Limitations and Ethics Statement}

\paragraph{Imperfect proxies.} The proxy functions we use to measure correctness are inherently imperfect. While they serve as reasonable approximations for distinguishing varying degrees of wrongness, they are not definitive, e.g. heuristic-based proxies may merely focus on the final answer and overlook the reasoning steps and model-based proxies are inaccurate and may suffer from domain shift problem. As a result, some wrong-over-wrong preferences may be misjudged, potentially affecting the evaluation of $\mathrm{Acc}_{\textit{WoW}}$ and analysis results.

\paragraph{Sensitivity to hyperparameters.} The alignment process we employ is sensitive to hyperparameter selection. This dependence can lead to variability in results across different models and datasets, making it challenging to guarantee consistent performance improvements. Further exploration of more robust hyperparameter configurations \citep{falkner2018bohb, arango2024quicktune} is necessary.

\paragraph{Limited scope of experimental datasets.} Experiments are conducted on datasets where ground-truth answers exist, and proxy functions can be employed to approximate wrongness. However, this setup doesn't fully reflect the application scenarios we aim to address — tasks where no clear ground-truth answers are available \citep{wang2024weaver}, or the problems are so challenging that even expert human annotators or LLMs might struggle \citep{welleck2021naturalproofs}. Evaluating the generalizability of our framework in these conditions remains an important avenue for future work.

\paragraph{Unique focus on knowledge and reasoning problems.} We focus on tasks and problems with absolute and indisputable correct answers in this work, such as multi-hop QA \citep{ding2023knowledge} and graph reasoning \citep{wang2023can}. Consequently, existing alignment works in these domains often assume access to correct answers and construct right-over-wrong pairs for preference learning \citep{wang2023making, cheng2024can, lin2024flame}. However, objective ``correct answers'' are often not feasible in general instruction following tasks, thus they employ human preference data between a pair of responses that is not necessarily right-over-wrong. Our scope is to investigate wrong-over-wrong alignment specifically focusing on the first type of knowledge/reasoning problems with absolute correctness while we leave general instruction following as future work.

%\section*{Ethics Statement}

\paragraph{Incomparable wrongness in social contexts.} In social contexts, wrongness is often subjective and potentially incomparable, which introduces biases and fairness concerns in wrong-over-wrong alignment. For instance, when judging the wrongness of statements related to sensitive topics such as politics or identity, cultural biases may influence the model's preferences \citep{xu2021detoxifying, bender2021dangers, feng2023pretraining}. A statement like ``People from \emph{certain} neighborhoods are more likely to commit crimes'' might be judged less wrong than ``All people from \emph{certain} neighborhoods are criminals'' but these evaluations reflect specific perspectives that may not be universally shared or appropriate in all contexts. Furthermore, aligning models with such judgments risks reinforcing harmful stereotypes or systemic biases. Ensuring fairness and transparency in these judgments is critical, and models should be designed to recognize when wrongness is subjective and abstain from making harmful comparisons.

\section*{Reproducibility Statement}

We provide all necessary details for the implementation and evaluation of our proposed wrong-over-wrong alignment approach in Appendix \ref{sec:config}. Specifically, Appendix \ref{sec:config} includes information of 8 datasets splits and preprocessing steps, 7 LLM checkpoints, hyperparameters selection and detailed configuration for each table. Moreover, the prompts for preference elicitation methods are explained in Table \ref{tab:prompt1} and Table \ref{tab:prompt2}. Our codes, including scripts for dataset generation, preference eliciting, preference accuracy evaluation, alignment, and evaluation of alignment are available publicly in repo \href{https://github.com/yaojh18/Varying-Shades-of-Wrong}{https://github.com/yaojh18/Varying-Shades-of-Wrong} to facilitate easy verification of our results.

\bibliography{iclr2025_conference}
\bibliographystyle{iclr2025_conference}

\newpage

\appendix

\section{Analysis (cont.)}

\begin{wraptable}{r}{0.5\linewidth}
    \centering
    % \scriptsize
    \setlength{\tabcolsep}{1pt}
    \vspace{-15pt}
    \resizebox{0.8\linewidth}{!}{
    \begin{tabular}{lcccccc}\toprule[1.5pt]
        Margin &KC &BG &COM$^2$ &NLG &overall \\
        \midrule[.75pt]\multicolumn{6}{c}{\textit{\ \ \textbf{Code-Alpaca}} }\\ \midrule[.75pt] 
        $M_{\textit{100}}$ &.183 &.258 &.221 &.174 &.220 \\
        $M_{\textit{50}}$ &.367 &.516 &.442 &.348 &.439 \\
        $M_{\textit{10}}$ &.425 &.544 &.466 &.482 &.505 \\
        \midrule[.75pt]\multicolumn{6}{c}{\textit{\ \ \textbf{FLAN}} }\\ \midrule[.75pt] 
        $M_{\textit{100}}$ &.000 &.098 &.000 &.096 &.080 \\
        $M_{\textit{50}}$ &.000 &.197 &.000 &.193 &.160 \\
        $M_{\textit{10}}$ &.000 &\textbf{.585} &.000 &\textbf{.610} &.487 \\
        \midrule[.75pt]\multicolumn{6}{c}{\textit{\ \ \textbf{Gemini-Alpaca}} }\\ \midrule[.75pt] 
        $M_{\textit{100}}$ &.297 &.356 &.396 &.347 &.349 \\
        $M_{\textit{50}}$ &.512 &.492 &.555 &.527 &.510 \\
        $M_{\textit{10}}$ &.534 &.458 &\textbf{.679} &.575 &.519 \\
        \midrule[.75pt]\multicolumn{6}{c}{\textit{\ \ \textbf{LIMA}} }\\ \midrule[.75pt]
        $M_{\textit{100}}$ &.110 &.079 &.215 &.099 &.098 \\
        $M_{\textit{50}}$ &.221 &.159 &.428 &.198 &.197 \\
        $M_{\textit{10}}$ &\textbf{.610} &.508 &.343 &.486 &.501 \\
        \midrule[.75pt]\multicolumn{6}{c}{\textit{\ \ \textbf{Open Assist}} }\\ \midrule[.75pt]
        $M_{\textit{100}}$ &.240 &.175 &.220 &.180 &.187 \\
        $M_{\textit{50}}$ &.481 &.351 &.441 &.360 &.375 \\
        $M_{\textit{10}}$ &.500 &.563 &.602 &.553 &\textbf{.555} \\
        \midrule[.75pt]\multicolumn{6}{c}{\textit{\ \ \textbf{Science}} }\\ \midrule[.75pt]
        $M_{\textit{100}}$ &.241 &.236 &.557 &.248 &.262 \\
        $M_{\textit{50}}$ &.482 &.472 &.561 &.482 &.482 \\
        $M_{\textit{10}}$ &.515 &.557 &.574 &.470 &.525 \\
        \midrule[.75pt]\multicolumn{6}{c}{\textit{\ \ \textbf{WizardLM}} }\\ \midrule[.75pt]
        $M_{\textit{100}}$ &.194 &.278 &.322 &.239 &.259 \\
        $M_{\textit{50}}$ &.388 &.508 &.474 &.478 &.482 \\
        $M_{\textit{10}}$ &.579 &.551 &.523 &.479 &.529 \\

        \bottomrule[1.5pt]
    \end{tabular}
    }
    \caption{
    Accuracy of wrong-over-wrong preference elicited from \textsc{Gemma-7B} fintuened on different instruction datasets. Finetuned LLMs have very different wrong-over-wrong preference accuracy across domains.
    }
    \label{tab:appendix_domain}
    \vspace{-40pt}
\end{wraptable}

\paragraph{More general LLMs as evaluators} \label{sec:more_llm} We present the rest of elicited wrong-over-wrong preferences, including evaluators being \textsc{Gemini-flash}, \textsc{Gemini-pro}, and \textsc{Mistral-7B}. The results in Table \ref{tab:appendix_acc_more} demonstrate that \textsc{GPT-4o} is still the best LLM at evaluating wrong-over-wrong preference and surpasses other LLMs by at least 12.0\% on overall $\mathrm{Acc}_{\textit{WoW}}$. We can also see leveraging \textsc{Gemini-flash}, \textsc{Gemini-pro} to evaluate answers generated by \textsc{GPT-4o} successfully mitigate the poor accuracy of self-evaluation.

\paragraph{Domain-specific LLMs as evaluators} To better figure out how $\mathrm{Acc}_{\textit{wow}}$ is relate to evaluators, we instruction finetune \textsc{Gemma-7B} on 7 different datasets \citep{ivison2023camels}, namely Code-Alpaca \citep{chaudhary2023code}, FLAN \citep{chung2024scaling}, Gemini-Alpaca, LIMA \citep{zhou2023lima}, Open Assistant 1 \citep{kopf2023openassistant}, Science literature and WizardLM Evol-Instruct V2 \citep{xu2023wizardlm}. Then we elicit wrong-over-wrong preference from them on Knowledge Crosswords, Bio Generation, COM$^2$ and NLGraph datasets with \textsc{GPT-3.5} being generator. We only experiment on score-based eliciting. Results in Table \ref{tab:appendix_domain} demonstrate that domain-specific LLMs as evaluators show considerable variance in their wrong-over-wrong preference elicitation accuracy across datasets. For example, \textsc{Gemma-7B} fintuned on FLAN achieves the best wrong-over-wrong accuracy on Bio Generation and NLGraph while unable to follow instructions on Knowledge Crosswords and COM$^2$. This suggests finetuning as a way to improve LLM-based wrong-over-wrong accuracy in certain domains.

\paragraph{More challenging datasets}
We investigate LLMs' capability to give wrong-over-wrong preferences on more challenging datasets, NLGraph (full) and ChessPuzzle (CP). NLGraph (full) employs all min/max questions in NLGraph, including very challenging questions like ``finding the maximum flow between two nodes in a weighted undirected graph'' and ``finding an assignment of jobs to applicants such that the maximum number of applicants find the job they are interested in''. Chess Puzzle contains chess puzzles from the \href{https://lichess.org/}{lichess.org} website \citep{lichess2023database}, where the inputs are board state represented in Forsyth–Edwards Notation (FEN) and LLMs are asked to generate the optimal next move for the current player. However, we find that LLMs tend to give invalid moves, which makes it hard to evaluate wrongness. Thus we employ Stockfish \citep{romstad2021stockfish}, a SOTA chess engine, to generate the 4 best moves for the current player and ask an LLM to choice from given options. Evaluators' task accuracy is 0.204 (random guess 0.000) on NLGraph (full) and 0.461 (random guess 0.250) on Chess Puzzle. The correctness proxy for NLGraph (full) is $p^{(\textit{nl)}}(\boldsymbol{a} \mid \boldsymbol{q}) = -|v_{\boldsymbol{a}} - v_{\textit{gt}}|$, where $v_{\boldsymbol{a}}$ is the extracted final answer and $v_{\textit{gt}}$ is the ground-truth answer. The correctness proxy for Chess Puzzle is $p^{(\textit{nl)}}(\boldsymbol{a} \mid \boldsymbol{q}) = wr_{\textit{Stockfish}} (\boldsymbol{a} \mid \boldsymbol{q})$, where $wr_{\textit{Stockfish}}$ is the win rate predicted by Stockfish. Experiment results in Table \ref{tab:appendix_dataset_more} demonstrate that the evaluator's $\mathrm{Acc}_{\textit{WoW}}$ drops significant on challenging tasks, e.g. $f_{\textit{GPT-4o}}^{(\textit{s})}$ with $M_{\textit{10}}$ drops by 23.7\%. The average $\mathrm{Acc}_{\textit{WoW}}$ across datasets for pairwise comparison with consistency check, logits-based with $M_{\textit{10}}$ and scored-based $M_{\textit{10}}$ are 0.585, 0.546, 0.550, indicating pairwise comparison may be more suitable for challenging tasks.

\begin{wraptable}{r}{0.5\linewidth}{
\centering
% \scriptsize
% \setlength{\tabcolsep}{1.2pt}
% \renewcommand{\arraystretch}{0.8}
\vspace{-15pt}
\resizebox{0.8\linewidth}{!}{
\begin{tabular}{lcccc}\toprule[1.5pt]
Margin &KC &NLG &Overall \\
\midrule[.75pt]\multicolumn{4}{c}{\textit{\ \ \textbf{batch size = 1}} }\\ \midrule[.75pt] 
$M_{\textit{100}}$ &0.447 &0.318 &0.383 \\
$M_{\textit{50}}$ &0.643 &0.596 &0.620 \\
$M_{\textit{10}}$ &0.719 &\textbf{0.701} &0.710 \\
\midrule[.75pt]\multicolumn{4}{c}{\textit{\ \ \textbf{batch size = 2}} }\\ \midrule[.75pt] 
$M_{\textit{100}}$ &0.482 &0.364 &0.423 \\
$M_{\textit{50}}$ &0.654 &0.575 &0.615 \\
$M_{\textit{10}}$ &0.708 &0.623 &0.666 \\
\midrule[.75pt]\multicolumn{4}{c}{\textit{\ \ \textbf{batch size = 5}} }\\ \midrule[.75pt] 
$M_{\textit{100}}$ &0.479 &0.383 &0.431 \\
$M_{\textit{50}}$ &0.733 &0.605 &0.669 \\
$M_{\textit{10}}$ &0.793 &0.652 &\textbf{0.723} \\
\midrule[.75pt]\multicolumn{4}{c}{\textit{\ \ \textbf{batch size = 10}} }\\ \midrule[.75pt] 
$M_{\textit{100}}$ &0.404 &0.299 &0.352 \\
$M_{\textit{50}}$ &0.723 &0.598 &0.661 \\
$M_{\textit{10}}$ &\textbf{0.804} &0.634 &0.719 \\
\bottomrule[1.5pt]
\end{tabular}}
\caption{Accuracy of scored-based wrong-over-wrong preference on Knowledge Crosswords and NLGraph (shortest path) dataset. The generator is \textsc{LLaMA3-8B} and the evaluator is \textsc{GPT-4o}. The number of answers generated to each question is 10. $b$ = 5 yields the best overall $\mathrm{Acc}_{\textit{WoW}}$. }\label{tab:appendix_batch}
}
\vspace{-40pt}
\end{wraptable}

\paragraph{Batch size for score-based eliciting}
We also investigate the effect of batch size $b$ on score-based preference eliciting. We employ \textsc{LLaMA3-8B} as generator and \textsc{GPT-4o} as evaluator. For simplicity, we only experiment on Knowledge Crosswords and NLGraph (shortest path) dataset. Experiment results in Table \ref{tab:appendix_batch} indicate that 5 is the best batch size for our experiment setup, which verifies our assumption that batch size has a trade-off effect on scored-based eliciting.

\paragraph{Consistency of pairwise comparison}
Pairwise comparison eliciting displays high sensitivity to positional bias. We report the proportions of consistent wrong-over-wrong preference after flipping for \textsc{LLaMA3-8B}, \textsc{GPT-3.5} and \textsc{GPT-4o} are 0.286, 0.128, 0.598. Stronger LLMs can generate more consistent preference, and small LLMs are not reliable evaluators.
\paragraph{Wrongness margin affecting $\mathrm{Acc}_{\textit{WoW}}$} \label{sec:wrong_margin} The accuracy of wrong-over-wrong judgement can also be affected by ``wrongness margin'' of a wrong-over-wrong pair. For example, a close-to-correct answer (e.g. ``the sun almost all rises in the east'') is obviously better than an absolutely wrong answer (e.g. ``the sun rises in the west''), while two wrong answers with close wrongness (e.g. ``the sun sometimes rises in the east'' and ``the sun sometimes perhaps rises in the east'') are hard to differentiate. We experiment on Knowledge Crosswords dataset and evaluate the wrongness margin is defined as the the absolute difference of correctly filled blanks. The results in Table \ref{tab:appendix_margin} demonstrate that a larger wrongness margin means easier evaluation, with average $\mathrm{Acc}_{\textsc{WoW}}$ = 0.696, 0.636 and 0.518 for wrongness margin being 3, 2, 1. This trend is particularly evident for highly wrong answer comparison, with 3 $\succ$ 2 having an accuracy of 0.533 while 1 $\succ$ 0 only has an accuracy of 0.463.

 \begin{wrapfigure}{r}{0.45\linewidth}
    \centering
    \vspace{-20pt}
    \includegraphics[width=0.98\linewidth]{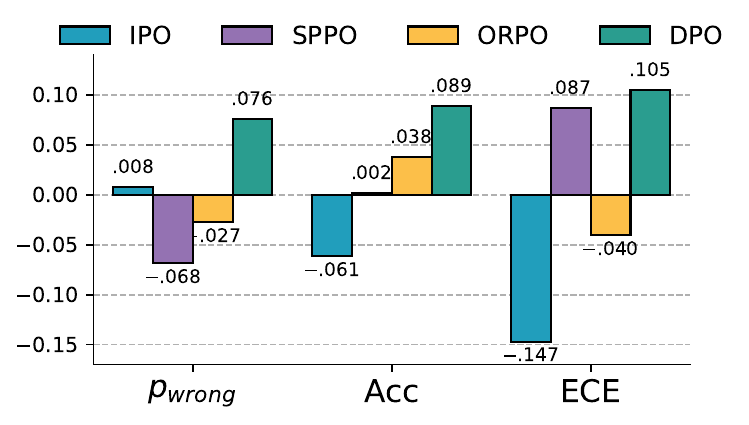}
    \caption{Evaluation different of preference optimization methods on less wrong, more correct and better calibration. The number is averaged over six experiment setups ((pairwise comparison with consistency check, score-based with $M_{\textit{10}}$, score-based with $M_{\textit{50}}$) $\times$ (Self-Generator, Mix-Generator)) on NLGraph dataset.}
    \vspace{-50pt}
    \label{fig:bar}
\end{wrapfigure}

\paragraph{No or negative effect of mixing generators is observed.} Experiments in Table \ref{tab:res_alignment} demonstrate no effect from mixing generators for $f_{\textit{GPT-4o}}^{(\textit{p})}$ with consistency checks, $f_{\textit{GPT-4o}}^{(\textit{s})}$ with $M_{\textit{10}}$ and oracle, with little change in average proxy scores discovered. And Table \ref{tab:res_alignment} also indicates an negative effect on $f_{\textit{GPT-4o}}^{(\textit{s})}$ with $M_{\textit{50}}$ with average improvement on $\Delta\mathrm{Acc}$ across datasets drops by 28.9\%. This suggests \emph{learning from the model's own mistakes is perhaps more important for wrong-over-wrong alignment}.

\paragraph{Preference Optimization Approach} To investigate the influence of different preference optimization methods, we also experiment wrong-over-wrong alignment with IPO \citep{azar2024general}, ORPO \citep{hong2024orpo} SimPO \citep{meng2024simpo} and SPPO \citep{wu2024self} on the NLGraph dataset. Results in Figure \ref{fig:bar} demonstrate that preference optimization methods other than DPO are sensitive to noise and could not be employed for wrong-over-wrong alignment without hyperparameter tuning. SimPO collapses on all 6 experiment setups, IPO collapses on 4 of them, and SPPO and ORPO show negative improvement compared to original \textsc{LLAMA3-8B}. This suggests wrong-over-wrong alignment could be noisy and we should adopt stable preference optimization methods like DPO \citep{xu2024dpo, chowdhury2024provably}. 

\begin{wraptable}{r}{0.5\linewidth}{
\centering
\vspace{-15pt}
% \scriptsize
% \setlength{\tabcolsep}{1.2pt}
% \renewcommand{\arraystretch}{0.8}
\resizebox{1\linewidth}{!}{
\begin{tabular}{lcccc}\toprule[1.5pt]
\multirow{3}{*}{\textbf{Method}} &\multicolumn{2}{c}{\textbf{MedMCQA}} &\multicolumn{2}{c}{\textbf{SciBench}} \\\cmidrule(lr){2-3}\cmidrule(lr){4-5}
&$\mathrm{Acc}_{\uparrow}$ &$\mathrm{ECE}_{\downarrow}$ &$\mathrm{Acc}_{\uparrow}$ &$\mathrm{ECE}_{\downarrow}$  \\
\midrule[0.75pt]
\textsc{Original} &0.566 &0.083 &0.105 &0.634 \\
\midrule[.75pt]
\multicolumn{5}{c}{\textit{\ \ \textbf{\textsc{Self-Generator}}} }\\ \midrule[.75pt]
\textsc{score} $M_{\textit{50}}$ &\textbf{0.565} &0.111 &0.095 &\cellcolor[HTML]{d4edda}0.607 \\
\textsc{score} $M_{\textit{10}}$ &0.550 &\cellcolor[HTML]{d4edda}\textbf{0.082} &0.068 &0.660 \\
\midrule[.75pt] \multicolumn{5}{c}{\textit{\ \ \textbf{\textsc{Mix-Generator}}} }\\ \midrule[.75pt]
\textsc{score} $M_{\textit{50}}$ &0.548 &0.083 &\cellcolor[HTML]{d4edda}\textbf{0.108} &\cellcolor[HTML]{d4edda}\textbf{0.591} \\
\textsc{score} $M_{\textit{10}}$ &0.553 &0.098 &0.095 &0.688 \\
\bottomrule[1.5pt]
\end{tabular}}
\caption{Generalization datasets without distinctive wrong-over-wrong separability. Wrong-over-wrong alignment is not helpful for domains without clear right-wrong distinctions.}
\label{tab:no_proxy}
}
\vspace{-5pt}
\end{wraptable}

\paragraph{Alignment on domains without clear right-wrong distinctions} \label{sec:no_proxy} We wonder if the effectiveness of wrong-over-wrong alignment is universal and doesn't rely on clear right-wrong distinctions within the task domain. To better prove this, we implement wrong-over-wrong alignment on SciBench \citep{wang2023scibench} and MedMCQA \citep{pal2022medmcqa} where no such proxy to evaluate the wrongness of answers exists. SciBench is a college-level scientific benchmark emphasizing scientific theorem application and numerical computation, while MedMCQA is a multiple-choice QA dataset in the medical domain. Experiment results in Table \ref{tab:no_proxy} indicate that wrong-over-wrong alignment is not universally helpful in domains without clear right-wrong distinctions. This suggests we should investigate more fine-grained separability among wrong answers and design a more sophisticated LLM-as-a-judge prompt.

\paragraph{Qualitative examples} \label{sec:example} We present qualitative examples demonstrating the improvement of wrong-over-wrong alignment on Knowledge Crosswords, Bio Generation, COM$^2$, NLGraph from Table \ref{tab:example11} to Table \ref{tab:example43}.

\section{Experiment Details} \label{sec:config}

\paragraph{Model Details} We employ 7 LLMs in the experiments, the detailed information of LLMs we used in this paper is as follows: 1) \textsc{LLaMA3-8B}, through the \href{https://huggingface.co/meta-llama/Meta-Llama-3-8B}{\textsc{meta-llama/Meta-Llama-3-8B}} checkpoint on Huggingface \citep{wolf2020transformers}; 2) \textsc{Mistral-7B}, through the \href{https://huggingface.co/mistralai/Mistral-7B-Instruct-v0.3}{\textsc{mistralai/Mistral-7B-Instruct-v0.3}} checkpoint on Huggingface; 3) \textsc{Gemma-7B}, we directly employ 7 instruction finetuned \textsc{Gemma-7B}s from \citet{ivison2023camels}; 4) \textsc{GPT-3.5}, through the \textsc{gpt-3.5-turbo-0125} checkpoint on OpenAI API; 5) \textsc{GPT-4o}, through the \textsc{gpt-4o-2024-05-13} checkpoint on OpenAI API; 6) \textsc{Gemini-Flash} through the \textsc{gemini-1.5-flash-001} by the Vertex AI API; 7) \textsc{Gemini-Pro} through the \textsc{gemini-1.5-pro-001} by the Vertex AI API.

\paragraph{Dataset Details} We employ 8 datasets in the experiments, 4 for the main experiments and 4 left in the appendix. The detailed information is:

\begin{itemize}[leftmargin=*]
    \item \textbf{Knowledge Crosswords (KC)} \citep{ding2023knowledge}: We downsample questions containing exactly three blanks, and we create four answer options by randomly choosing 0, 1, 2, or 3 correctly filled blanks. We sample 625 questions from the official dataset and split them into train, validation, test sets with an 8:1:1 ratio.
    \item \textbf{NLGraph (NLG)} \citep{wang2023can, zhang2024can} We employ the ``shortest path'' subset for main experiments denoted as NLGraph (shortest path) and ``shortest path'', ``maximum flow'' and ``matching'' subsets for experiments in the appendix denoted as NLGraph (full). Correctness proxy for ``maximum flow'' and ``matching'' subset is: $p^{(\textit{mf)}}(\boldsymbol{a} \mid \boldsymbol{q}) = 1-\frac{|f_{\textit{gt}} - f_{\boldsymbol{a}}|}{f_{\textit{gt}}}$, with $f_{\boldsymbol{a}}$ being value given in $\boldsymbol{a}$ and $f_{\textit{gt}}$ being the ground-truth. We sample 380, 350 and 600 questions in ``shortest path'', ``maximum flow'' and ``matching'' subsets and split them into train, validation, and test sets with an 8:1:1 ratio.
    \item \textbf{Bio Generation (BG)} We employ the bio generation dataset and retrieved documents provided in FActScore \citep{min2023factscore}. We sample 625 questions from the official dataset and split them into train, validation, test sets with an 8:1:1 ratio.
    \item \textbf{COM$^2$} \citep{fang2024complex}: We employ the ``2i'' and ``3i'' question subsets of COM$^2$, where we empirically find that Vera could generate more accurate scores. We sample 625 questions from the official dataset and split them into train, validation, test sets with an 8:1:1 ratio.
    \item \textbf{Hellaswag} \citep{zellers2019hellaswag}: Correctness proxy function for HellaSwag is: $p^{(\textit{(hs)}}(\boldsymbol{a} \mid \boldsymbol{q}) = \mathrm{Vera}(\boldsymbol{a} \mid \boldsymbol{q}) \in [0,1]$. We sample 125 questions from the official validation split and split them into validation, test sets with a 1:1 ratio.
    \item \textbf{Chess Puzzle} \citep{lichess2023database}: We downsample the beginner-level chess puzzles with ELO ranging from 600 to 1000. We sample 625 questions from the official dataset and split them into train, validation, test sets with an 8:1:1 ratio.
    \item \textbf{SciBench} \citep{wang2023scibench}: We sample 625 questions from the official dataset and split them into train, validation, and test sets with an 8:1:1 ratio.
    \item \textbf{MedMCQA} \citep{pal2022medmcqa}: We sample 500 questions from the official train split as train set. We sample 125 questions from the official validation split into validation, test sets with a 1:1 ratio.
\end{itemize}

\paragraph{Implementation Details}
% [details of what we did as our method] 1) part 1 details elicitation methods 2) tuning hyperparameters
\begin{itemize}[leftmargin=*]
    \item \textbf{Pairwise comparison}. The exact prompt we use for pairwise comparison is in Table \ref{tab:prompt1}.
    \item \textbf{Score-based} The exact prompt we use for score-based eliciting is in Table \ref{tab:prompt2}.
    \item \textbf{Generation of wrong-over-wrong dataset} The pipeline used to generate $\mathcal{D}_{\textit{WoW}}$ is described in Algorithm \ref{alg:WoW_generation}.
    \item \textbf{QLora finetuning} We employ \href{https://github.com/unslothai/unsloth}{Unsloth} and Transformers libraries for preference optimization. We apply grid search on learning rate (1e-4, 5e-5, 1e-5), learning rate scheduler (cosine, cosine with restart and reduce lr on plateau), weight decay (0, 1e-5, 1e-3) and number of train epochs (1, 3, 5) for main experiments and right-over-wrong alignment experiments. We use random seed = 42 for all of our experiments. 
\end{itemize}

\begin{figure}[p]
    \centering
    \begin{minipage}{0.45\linewidth}
        \centering
        \includegraphics[width=0.9\linewidth]{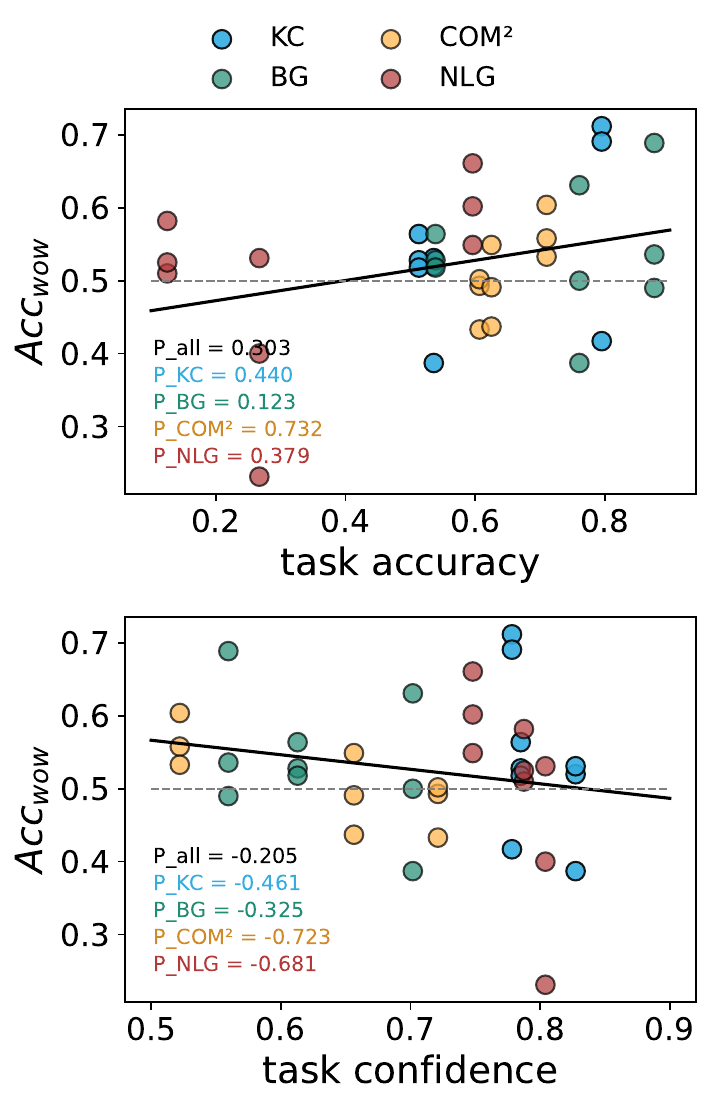}
        \caption{Correlation between task accuracy, confidence and $\mathrm{Acc}_{\textit{WoW}}$ for pairwise comparison with consistency check. Data points are from all 3 LLMs we used to elicit wong-over-wrong preferences. $P$ stands for Pearson correlation coefficient.}
        \label{fig:corr1_pairwise}
    \end{minipage}
    \hspace{0.05\linewidth}
    \begin{minipage}{0.45\linewidth}
        \centering
        \includegraphics[width=0.9\linewidth]{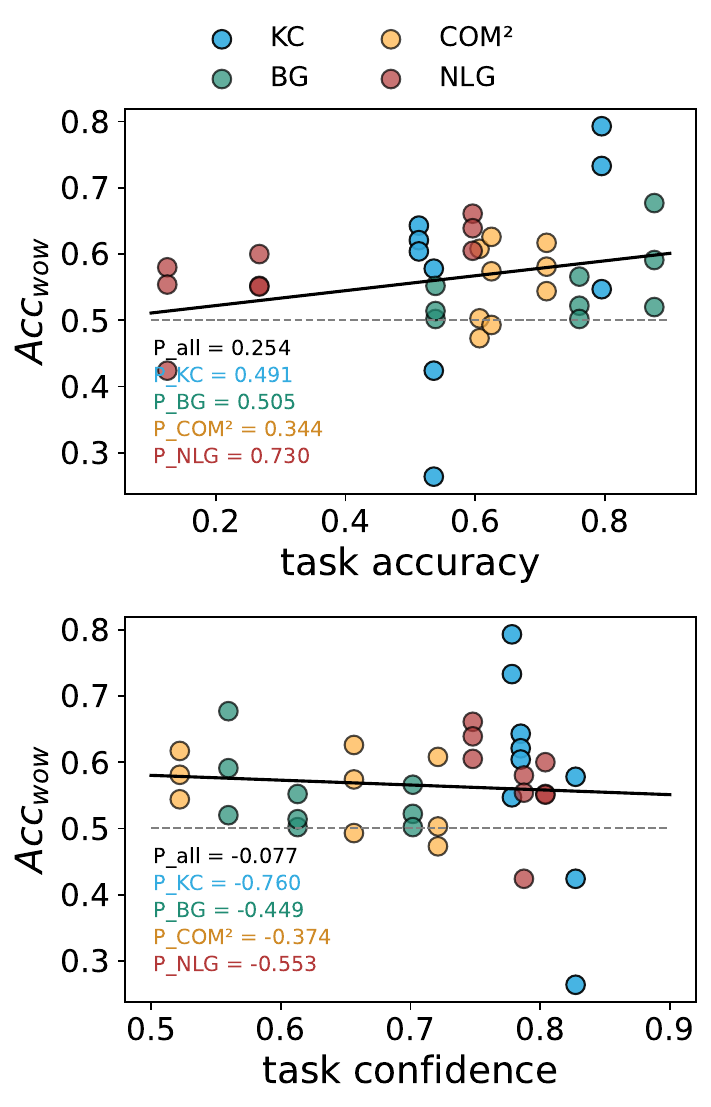}
        \caption{Correlation between task accuracy, confidence and $\mathrm{Acc}_{\textit{WoW}}$ for score-based with $M_{\textit{50}}$. Data points are from all 3 LLMs we used to elicit wrong-over-wrong preferences. $P$ stands for Pearson correlation coefficient.}
        \label{fig:corr1_score}
    \end{minipage}
\end{figure}

\paragraph{Experiment Details} In Table \ref{tab:res_preference}, we use batch size = 5 for all score methods due to optimal empirical results. In Table \ref{tab:res_alignment}, we define FActScore \textgreater \ 0.9 as ``correct'' on Bio Generation dataset; all the preferences are elicited from \textsc{GPT-4o}. In Table \ref{tab:ana_row} the best wrong-over-wrong alignment setup for Knowledge Crosswords, Bio Generation, COM$^2$ and NLGraph are score-based with $M_{\textit{50}}$ on self-generated data, pairwise comparison with consistency check on mix-generated data and score-based with $M_{\textit{10}}$ on self-generated data; Bio Generation dataset is omitted because there is not well-defined ground-truth on this dataset. Correlations between task accuracy, confidence and other wrong-over-wrong preference eliciting methods are displayed in Figure \ref{fig:corr1_pairwise} and Figure \ref{fig:corr1_score}. 

\begin{table}[p]\centering
% \scriptsize
\setlength{\tabcolsep}{1.5pt}
\resizebox{0.9\textwidth}{!}{
\begin{tabular}{llcccc|cccc|cccc|c}\toprule[1.5pt]
\multirow{2}{*}{\textbf{Method}}& \multirow{2}{*}{\textbf{Margin}}&\multicolumn{4}{c}{\textsc{LLaMA3-8B}} &\multicolumn{4}{c}{\textsc{GPT-3.5}} &\multicolumn{4}{c}{\textsc{GPT-4o}} &\multirow{3}{*}{Overall} \\
% \cmidrule{3-14}
 & &KC &BG &COM$^2$ &NLG &KC &BG &COM$^2$ &NLG &KC &BG &COM$^2$ &NLG & \\

\midrule[.75pt]\multicolumn{15}{c}{\textit{\ \ \textbf{\textsc{Model-independent}}} }\\ \midrule[.75pt] 
\textsc{Heuristic} &$M_{\textit{100}}$ &.491 &.486 &.488 &.498 &.519 &.504 &.473 &.466 &.547 &.471 &.473 &.502 & .489\\
\textsc{Consistency} &$M_{\textit{100}}$ &.296 &- &.287 &.360 &.318 &- &.279 &.336 &.311 &- &.325 &.313 & .328\\

\midrule[.75pt]\multicolumn{15}{c}{\textit{\ \ \textbf{\textsc{LLaMA3-8B} as Evaluator}} }\\ \midrule[.75pt] 
\textsc{Logits}&$M_{\textit{100}}$ &.527 &.540 &.525 &.508 &.546 &.516 &.506 &.477 & .528 & .491 &.440 & .546 & .519\\
\textsc{Score}&$M_{\textit{100}}$ &.479 &.386 &.385 &.415 &.496 &.376 &.48 &.434 &.387 &.373 &.525 &.273 & .400\\

\midrule[.75pt]\multicolumn{15}{c}{\textit{\ \ \textbf{\textsc{GPT-3.5} as Evaluator} }}\\ \midrule[.75pt] 
\textsc{Logits}&$M_{\textit{100}}$ &- &- &- &- &.528 &.508 &.552 &.479 &- &- &- &- & .504\\
\textsc{Score} &$M_{\textit{100}}$ &.212 &.352 &.287 &.323 &.289 &.327 &.458 &.422 &.132 &.273 &.269 &.307 &.335\\

\midrule[.75pt]\multicolumn{15}{c}{\textit{\ \ \textbf{\textsc{GPT-4o} as Evaluator}} }\\ \midrule[.75pt] 
\textsc{Logits}&$M_{\textit{100}}$ &- &- &- &- &- &- &- &- &.472 &.526 &.478 &.540 & .526\\
\multirow{2}{*}{\textsc{Pairwise}} &all &.605 &.593 &.507 &.551 &.646 &.512 &.515 &.577 &.434 &.501 &.526 &.537 & .562\\
&filter &.691 &.689 &.533 &.602 &.712 &.536 &.558 &.661 &.417 &.490 &.604 &.549 & .624\\
\multirow{3}{*}{\textsc{Score}}&$M_{\textit{100}}$ &.479 &.385 &.450 &.383 &.540 &.351 &.484 &.408 &.481 &.353 &.467 &.369 & .393\\
&$M_{\textit{50}}$ &.733 &.677 &.544 &\underline{.605} &\underline{.793} &\underline{.591} &.617 &.661 &.547 &.520 &.581 &\textbf{.639} & \underline{.641}\\
&$M_{\textit{10}}$ &\textbf{.793} &\textbf{.795} &.534 &\textbf{.652} &\textbf{.835} &\textbf{.655} &\underline{.711} &\textbf{.684} &.400 &.586 &.520 &.578 & \textbf{.709}\\

\midrule[.75pt]\multicolumn{15}{c}{\textit{\ \ \textbf{\textsc{Mistral-7B} as Evaluator}} }\\ \midrule[.75pt] 
\multirow{2}{*}{\textsc{Pairwise}} & all &.463 &.471 &.462 &.492 &.447 &.482 &.524 &.501 &.453 &.503 &.471 &.514 &.482 \\
&filter &.551 &.427 &.451 &.505 &.531 &.491 &.574 &.538 &.542 &.365 &.542 &.560 &.514 \\
\multirow{3}{*}{\textsc{Score}}&$M_{\textit{100}}$ &.351 &.356 &.370 &.389 &.379 &.348 &.431 &.453 &.051 &.305 &.354 &.337 &.366 \\
&$M_{\textit{50}}$ &.641 &.538 &.505 &.527 &.652 &.512 &.584 &.599 &.102 &.514 &.632 &.576 &.549 \\
&$M_{\textit{10}}$ &.683 &.558 &.486 &.558 &.723 &.513 &.626 &.565 &.222 &.557 &.677 &.565 &.565 \\

\midrule[.75pt]\multicolumn{15}{c}{\textit{\ \ \textbf{\textsc{Gemini-flash} as Evaluator}} }\\ \midrule[.75pt] 
\multirow{2}{*}{\textsc{Pairwise}} & all &.531 &.505 &.500 &.509 &.519 &.473 &.538 &.574 &.415 &.471 &.519 &.531 &.509\\
&filter &.574 &.540 &.517 &.520 &.557 &.444 &.561 &.629 &.410 &.373 &.556 &.565 &.529\\
\multirow{3}{*}{\textsc{Score}}&$M_{\textit{100}}$ &.482 &.433 &.433 &.362 &.521 &.368 &.497 &.455 &.453 &.377 &.560 &.429 &.420\\
&$M_{\textit{50}}$ & .642 &.584 &.510 &.517 &.661 &.497 &.630 &.616 &.528 &.510 &.674 &\underline{.600} & .567\\
&$M_{\textit{10}}$ &\underline{.746} &.606 &.504 &.551 &.719 &.483 &\textbf{.756} &\underline{.665} &\textbf{.800} &\textbf{.593} &\textbf{.817} &.576 & .599\\

\midrule[.75pt]\multicolumn{15}{c}{\textit{\ \ \textbf{\textsc{Gemini-pro} as Evaluator}} }\\ \midrule[.75pt] 
\multirow{2}{*}{\textsc{Pairwise}} &all &.539 &.538 &.493 &.513 &.558 &.435 &.409 &.554 &.396 &.437 &.522 &.530 & .509\\
&filter &.598 &.588 &.547 &.564 &.628 &.471 &.557 &.658 &.423 &.412 &.560 &.569 & .557\\
\multirow{3}{*}{\textsc{Score}}&$M_{\textit{100}}$ &.494 &.442 &.450 &.397 &.555 &.361 &.483 &.476 &.434 &.363 &.460 &.417 & .428\\
&$M_{\textit{50}}$ &.653 &.612 &\underline{.551} &.520 &.692 &.520 &.600 &.631 &.566 &.511 &.567 &.560 & .581\\
&$M_{\textit{10}}$ &.742 &\underline{.695} &\textbf{.573} &.533 &.762 &.524 &.681 &.662 &\underline{.600} &\underline{.590} &\underline{.732} &.572 & .633\\

\bottomrule[1.5pt]
\end{tabular}
}
\caption{Accuracy of wrong-over-wrong preference elicited from \textsc{Mistral-7B}, \textsc{Gemini-flash} and \textsc{Gemini-pro}, and also the accuracy of heuristic, consistency-based, logits-based and score-based methods with a margin of $M_{\textit{100}}$ that are not displayed in the main paper. $f_{\textit{GPT-4o}}^{(\textit{s})}$ with $M_{\textit{10}}$ still achieves the best weighted average $\mathrm{Acc}_{\textit{WoW}}$ = 0.709 across all datasets.}
\label{tab:appendix_acc_more}
\end{table}

\begin{table}[p]
\centering
% \scriptsize
% \setlength{\tabcolsep}{1pt}
\resizebox{0.8\textwidth}{!}{
\begin{tabular}{llcc|cc|cc|c} \toprule[1.5pt]
\multirow{2}{*}{\textbf{Method}} & \multirow{2}{*}{\textbf{Margin}} & \multicolumn{2}{c}{\textsc{LLaMA3-8B}} & \multicolumn{2}{c}{\textsc{GPT-3.5}} & \multicolumn{2}{c}{\textsc{GPT-4o}} & \multirow{2}{*}{Overall} \\
 &  & NLG & CP & NLG & CP & NLG & CP & \\
\midrule[0.75pt]
\multicolumn{9}{c}{\textit{\textbf{Model-independent}}} \\ \midrule[0.75pt]
\multirow{3}{*}{\textsc{Heuristic}} & $M_{\textit{100}}$ & 0.510 & 0.506 & 0.472 & 0.504 & 0.535 & 0.481 & 0.501 \\
 & $M_{\textit{50}}$ & 0.509 & 0.506 & 0.468 & 0.519 & 0.547 & 0.464 & 0.502 \\
 & $M_{\textit{10}}$ & 0.457 & 0.512 & 0.476 & 0.501 & 0.486 & 0.475 & 0.478 \\
\multirow{3}{*}{\textsc{Consistency}} & $M_{\textit{100}}$ & 0.359 & 0.281 & 0.323 & 0.300 & 0.269 & 0.301 & 0.318 \\
 & $M_{\textit{50}}$ & 0.544 & 0.478 & 0.514 & 0.499 & 0.432 & 0.498 & 0.505 \\
 & $M_{\textit{10}}$ & 0.509 & 0.535 & 0.506 & 0.492 & 0.302 & 0.468 & 0.478 \\
\midrule[0.75pt]
\multicolumn{9}{c}{\textit{\textbf{LLaMA3-8B as Evaluator}}} \\ \midrule[0.75pt]
\multirow{2}{*}{\textsc{Pairwise}} & all & 0.501 & 0.516 & 0.477 & 0.493 & 0.551 & 0.504 & 0.503 \\
 & filter & 0.586 & 0.560 & 0.538 & 0.504 & 0.500 & 0.510 & 0.545 \\
\multirow{3}{*}{\textsc{Logits}} & $M_{\textit{100}}$ & 0.527 & 0.495 & 0.469 & 0.507 & 0.574 & 0.525 & 0.511 \\
 & $M_{\textit{50}}$ & 0.539 & 0.488 & 0.456 & 0.508 & 0.625 & 0.532 & 0.518 \\
 & $M_{\textit{10}}$ & 0.562 & 0.512 & 0.453 & 0.485 & 0.656 & 0.525 & 0.530 \\
\multirow{3}{*}{\textsc{Score}} & $M_{\textit{100}}$ & 0.426 & 0.411 & 0.421 & 0.406 & 0.412 & 0.372 & 0.417 \\
 & $M_{\textit{50}}$ & 0.547 & 0.524 & 0.536 & 0.534 & 0.593 & 0.529 & 0.546 \\
 & $M_{\textit{10}}$ & 0.529 & \underline{0.564} & 0.552 & \underline{0.545} & \underline{0.636} & 0.511 & 0.557 \\
\midrule[0.75pt]
\multicolumn{9}{c}{\textit{\textbf{GPT-3.5 as Evaluator}}} \\ \midrule[0.75pt]
\multirow{2}{*}{\textsc{Pairwise}} & all & 0.505 & 0.485 & 0.514 & 0.469 & 0.507 & 0.499 & 0.502 \\
 & filter & \textbf{0.802} & 0.547 & \textbf{0.590} & \textbf{0.585} & \textbf{0.882} & 0.535 & \textbf{0.642} \\
\multirow{3}{*}{\textsc{Logits}} & $M_{\textit{100}}$ & - & - & 0.479 & 0.518 & - & - & 0.489 \\
 & $M_{\textit{50}}$ & - & - & 0.472 & 0.521 & - & - & 0.484 \\
 & $M_{\textit{10}}$ & - & - & 0.492 & 0.515 & - & - & 0.498 \\
\multirow{3}{*}{\textsc{Score}} & $M_{\textit{100}}$ & 0.368 & 0.311 & 0.401 & 0.355 & 0.333 & 0.219 & 0.359 \\
 & $M_{\textit{50}}$ & 0.579 & 0.510 & 0.540 & 0.538 & 0.593 & 0.438 & 0.553 \\
 & $M_{\textit{10}}$ & 0.578 & 0.496 & 0.527 & 0.501 & 0.614 & \textbf{0.635} & 0.553 \\
\midrule[0.75pt]
\multicolumn{9}{c}{\textit{\textbf{GPT-4o as Evaluator}}} \\ \midrule[0.75pt]
\multirow{2}{*}{\textsc{Pairwise}} & all & 0.560 & 0.495 & 0.554 & 0.512 & 0.575 & 0.503 & 0.546 \\
 & filter & \underline{0.586} & 0.519 & \underline{0.569} & 0.520 & 0.613 & 0.512 & 0.568 \\
\multirow{3}{*}{\textsc{Logits}} & $M_{\textit{100}}$ & - & - & - & - & 0.546 & 0.525 & 0.564 \\
 & $M_{\textit{50}}$ & - & - & - & - & 0.577 & \underline{0.532} & 0.594 \\
 & $M_{\textit{10}}$ & - & - & - & - & 0.630 & 0.525 & \underline{0.617} \\
\multirow{3}{*}{\textsc{Score}} & $M_{\textit{100}}$ & 0.352 & 0.346 & 0.344 & 0.402 & 0.399 & 0.381 & 0.361 \\
 & $M_{\textit{50}}$ & 0.573 & 0.523 & 0.538 & 0.513 & 0.554 & 0.487 & 0.546 \\
 & $M_{\textit{10}}$ & 0.551 & \textbf{0.611} & 0.535 & 0.490 & 0.529 & 0.397 & 0.541 \\
\bottomrule[1.5pt]
\end{tabular}
}
\caption{Accuracy of wrong-over-wrong preference on NLGraph (full) and ChessPuzzle dataset. $f_{\textit{GPT-3.5}}^{(\textit{p})}$ with consistency check achieve the best weighted average $\mathrm{Acc}_{\textit{WoW}}$ = 0.642 across all datasets.}
\label{tab:appendix_dataset_more}
\end{table}

\begin{table}[p]
\centering
% \scriptsize
\resizebox{0.6\textwidth}{!}{
\begin{tabular}{lccccccc}\toprule[1.5pt]
Margin &3 $\succ$ 0 &3 $\succ$ 1 &2 $\succ$ 0 &3 $\succ$ 2 &2 $\succ$ 1 &1 $\succ$ 0 \\
\midrule[.75pt]\multicolumn{7}{c}{\textit{\ \ \textbf{logits (self)}} }\\ \midrule[.75pt]
$M_{\textit{100}}$ &\textbf{0.571} &0.543 &0.557 &0.500 &0.520 &0.502 \\
$M_{\textit{50}}$ &\textbf{0.590} &0.553 &0.551 &0.514 &0.533 &0.531 \\
$M_{\textit{10}}$ &\textbf{0.649} &0.576 &0.625 &0.540 &0.464 &0.629 \\
\midrule[.75pt]\multicolumn{7}{c}{\textit{\ \ \textbf{score (LLaMA3-8B)}} }\\ \midrule[.75pt]
$M_{\textit{100}}$ &\textbf{0.574} &0.527 &0.512 &0.423 &0.467 &0.459 \\
$M_{\textit{50}}$ &\textbf{0.749} &0.711 &0.658 &0.541 &0.610 &0.587 \\
$M_{\textit{10}}$ &0.758 &\textbf{0.774} &0.714 &0.607 &0.667 &0.645 \\
\midrule[.75pt]\multicolumn{7}{c}{\textit{\ \ \textbf{score (GPT-3.5)}} }\\ \midrule[.75pt]
$M_{\textit{100}}$ &\textbf{0.308} &0.238 &0.233 &0.189 &0.188 &0.236 \\
$M_{\textit{50}}$ &\textbf{0.463} &0.454 &\textbf{0.463} &0.399 &0.392 &0.431 \\
$M_{\textit{10}}$ &\textbf{0.780} &0.776 &0.670 &0.617 &0.585 &0.434 \\
\midrule[.75pt]\multicolumn{7}{c}{\textit{\ \ \textbf{score (GPT-4o)}} }\\ \midrule[.75pt]
$M_{\textit{100}}$ &\textbf{0.753} &0.668 &0.575 &0.509 &0.481 &0.339 \\
$M_{\textit{50}}$ &\textbf{0.917} &0.895 &0.812 &0.724 &0.764 &0.508 \\
$M_{\textit{10}}$ &\textbf{0.958} &0.940 &0.829 &0.792 &0.867 &0.529 \\\midrule[.75pt]
average&\textbf{0.696} &0.665 &0.607 &0.533 &0.558 &0.463 \\
\bottomrule[1.5pt]
\end{tabular}
}
\caption{Preference accuracy for wrong-over-wrong pair with different wrongness margins. $a \succ b$ means preferring an answer filling in $a$ blanks correctly than an answer filling in $b$ blanks correctly. To have an overview, we also include correct answers which fill in 3 blanks correctly. We experiment on score-based eliciting with 3 LLM evaluators respectively and logits-based with self-evaluation. The results are an average on 3 generators. The largest wrongness margin 3 $\succ$ 1 yields the most accurate preference.}\label{tab:appendix_margin}
\end{table}

\begin{table}[p]
\centering
% \scriptsize
\begin{tabular}{p{0.95\linewidth}}
\toprule[1.5pt]
\textbf{Prompt:} You are a helpful assistant whose goal is to select the preferred AI model's output for a given instruction based on their correctness. \\
\textbf{Instruction:} \\
\texttt{[question here]} \\
\textbf{Output 1:} \\
\texttt{[answer $\boldsymbol{a}_1$ here]} \\
\textbf{Output 2:} \\
\texttt{[answer $\boldsymbol{a}_2$ here]} \\
Please first briefly describe your reasoning (in less than 100 words), and then conclude with ``Preferred output: 1 or 2.'' in the last line. Let’s think step-by-step. \\
\bottomrule[1.5pt]
\end{tabular}
\caption{Prompt for pairwise comparison wrong-over-wrong preference eliciting.} \label{tab:prompt1}
\end{table}

\begin{table}[p]
\centering
% \scriptsize
\begin{tabular}{p{0.95\linewidth}}
\toprule[1.5pt]
\textbf{Prompt:} Below is an instruction from a user and five candidate responses. Please grade the correctness of each response on a 5-point scale from 0 to 5, where 0 is the lowest correctness, and 5 is the highest correctness. Only use integers for the ratings. \\
\textbf{Instruction:} \\
\texttt{[question here]} \\
\textbf{Response 1:} \\
\texttt{[answer $\boldsymbol{a}_1$ here]} \\
\textbf{Response 2:} \\
\texttt{[answer $\boldsymbol{a}_2$ here]} \\
\textbf{Response 3:} \\
\texttt{[answer $\boldsymbol{a}_3$ here]} \\
\textbf{Response 4:} \\
\texttt{[answer $\boldsymbol{a}_4$ here]} \\
\textbf{Response 5:} \\
\texttt{[answer $\boldsymbol{a}_5$ here]} \\
Please first briefly describe your reasoning (in less than 100 words for each response), followed by a new line in the format of ``Score: \textless rating\textgreater''. Let’s think step-by-step. \\
\bottomrule[1.5pt]
\end{tabular}
\caption{Prompt for score-based wrong-over-wrong preference eliciting.} \label{tab:prompt2}
\end{table}

\begin{algorithm}[p]
\caption{$\mathcal{D}_{\textit{WoW}}$ generation pipeline}
\begin{algorithmic} \label{alg:WoW_generation}
    \STATE \textbf{Input:} Original dataset $\mathcal{D}_{0} = \{(\boldsymbol{q}^{\textit{(i)}}, \boldsymbol{a}_{gt}^{\textit{(i)}})\}_{i=1}^n$
    \STATE $\mathcal{D}_{\textit{WoW}} = \emptyset$
    \FOR{$i$ in $1, 2, \dots, n$}
        \STATE Sample $m$ answers from LLM: $\{\boldsymbol{a}_j^{\textit{(i)}}\}_{j=1}^m \sim \mathrm{LLM}(\boldsymbol{q}^{\textit{(i)}})$\
        \STATE Filter out correct answers: $\{\boldsymbol{a}_j^{\textit{(i)}}\mid \boldsymbol{a}_{j}^{\textit{(i)}} \neq \boldsymbol{a}_{gt}^{\textit{(i)}}\}_{j=1}^{m'}$
        \FOR{$\boldsymbol{a}_j^{\textit{(i)}}$, $\boldsymbol{a}_k^{\textit{(i)}}$ in $\mathrm{combination}(\{\boldsymbol{a}_j^{\textit{(i)}}\}_{j=1}^{m'})$}
            \IF{$f_{\textit{LLM}}(\boldsymbol{a}_j^{\textit{(i)}} \succ \boldsymbol{a}_k^{\textit{(i)}} \mid \boldsymbol{q}^{\textit{(i)}}) = 1$}
                \STATE $\boldsymbol{a}_c,\boldsymbol{a}_r = \boldsymbol{a}_j^{\textit{(i)}}, \boldsymbol{a}_k^{\textit{(i)}}$
            \ELSIF{$f_{\textit{LLM}}(\boldsymbol{a}_j^{\textit{(i)}} \succ \boldsymbol{a}_k^{\textit{(i)}} \mid \boldsymbol{q}^{\textit{(i)}}) = -1$}
                \STATE $\boldsymbol{a}_c,\boldsymbol{a}_r = \boldsymbol{a}_k^{\textit{(i)}}, \boldsymbol{a}_j^{\textit{(i)}}$
            \ENDIF
            \STATE Append ($\boldsymbol{q}^{\textit{(i)}}, \boldsymbol{a}_c,\boldsymbol{a}_r$) to $\mathcal{D}_{\textit{WoW}}$ 
        \ENDFOR
    \ENDFOR
    \RETURN $\mathcal{D}_{\textit{WoW}}$
\end{algorithmic}
\end{algorithm}

\begin{table}[t]
\centering
% \scriptsize
\begin{tabular}{p{0.95\linewidth}}
\toprule[1.5pt]

\textbf{Question:}

Instruction: Pick the correct answer for each blank that satisfies all the given constraints.  
Constraints: (Miles\_Goodman, wroteMusicFor, blank 3); (Julien\_Temple, directed, blank 1); (blank 2, actedIn, blank 3); (blank 2, actedIn, blank 1); (Jason\_Alexander, actedIn, blank 3);  

\\
\textbf{Correct answer:}

blank 1: Earth\_Girls\_Are\_Easy, blank 2: Damon\_Wayans, blank 3: Blankman  

\\
\textbf{Original answer:}  

Let's analyze the question step by step:  
1. Julien Temple directed blank 1. From the options, we can conclude that blank 1 is either "Earth\_Girls\_Are\_Easy" or "Burn\_'Em\_Up\_Barnes".  
2. Miles Goodman wrote music for blank 3. From the options, we can conclude that blank 3 is either "Blankman" or "On\_Guard\_(1997\_film)".  
3. blank 2 acted in blank 1 and blank 3. From the options, we can conclude that blank 2 is the same person in both cases. The only options that satisfy this constraint are B and D.  
4. Jason Alexander acted in blank 3. From the options, we can conclude that blank 3 is either "Blankman" or "On\_Guard\_(1997\_film)". Given the previous conclusion, blank 3 must be "Blankman".  
5. Putting it all together, we can conclude that the correct answer is B. blank 1: Burn\_'Em\_Up\_Barnes, blank 2: Justin\_Theroux, blank 3: Blankman.  

Final answer: B.  

\\
\textbf{Less wrong answer:}  

Let's analyze the question step by step:  
1. Julien\_Temple, directed, blank 1: Julien Temple directed Earth\_Girls\_Are\_Easy, so blank 1 must be Earth\_Girls\_Are\_Easy.  
2. blank 2, actedIn, blank 1: From step 1, blank 1 is Earth\_Girls\_Are\_Easy. So, blank 2 must have acted in Earth\_Girls\_Are\_Easy. Let's call blank 2 X.  
3. blank 2, actedIn, blank 3: From step 2, X acted in Earth\_Girls\_Are\_Easy and also acted in blank 3.  
4. Jason\_Alexander, actedIn, blank 3: Jason Alexander acted in blank 3.  
5. blank 2, actedIn, blank 1: From step 2, X acted in Earth\_Girls\_Are\_Easy again (because it's the same film).  
6. Miles\_Goodman, wroteMusicFor, blank 3: Miles Goodman wrote music for blank 3.  
7. Julien\_Temple, directed, blank 1: We already knew this, but it's repeated to ensure consistency.  

Now, let's analyze the options:  
A. blank 1: Burn\_'Em\_Up\_Barnes, blank 2: Justin\_Theroux, blank 3: On\_Guard\_(1997\_film): Burn\_'Em\_Up\_Barnes is not directed by Julien Temple, and Justin Theroux didn't act in On\_Guard\_(1997\_film). Discarded.  
B. blank 1: Earth\_Girls\_Are\_Easy, blank 2: Damon\_Wayans, blank 3: Blankman: We already know that blank 1 is Earth\_Girls\_Are\_Easy. Damon Wayans didn't act in Earth\_Girls\_Are\_Easy, and Blankman isn't the same film as On\_Guard\_(1997\_film). Discarded.  
C. blank 1: Burn\_'Em\_Up\_Barnes, blank 2: Justin\_Theroux, blank 3: Blankman: Burn\_'Em\_Up\_Barnes is not directed by Julien Temple, and Justin Theroux didn't act in Blankman. Discarded.  
D. blank 1: Earth\_Girls\_Are\_Easy, blank 2: Justin\_Theroux, blank 3: Blankman: This option fits all the constraints:  
- blank 1 is Earth\_Girls\_Are\_Easy (step 1)  
- blank 2 is Justin\_Theroux (step 2)  
- blank 3 is Blankman (works with steps 3, 4, and 6)  

So, the final answer is:  
D. blank 1: Earth\_Girls\_Are\_Easy, blank 2: Justin\_Theroux, blank 3: Blankman  

\textcolor{red}{(Correctly filled blanks 1 $\rightarrow$ 2) } 

\\
\textbf{More correct answer:}  

Let's break down the constraints step by step:  
1. (Miles\_Goodman, wroteMusicFor, blank 3)  
This constraint tells us that Miles Goodman wrote music for a movie, and the title of that movie is blank 3.  
2. (Julien\_Temple, directed, blank 1)  
This constraint tells us that Julien Temple directed a movie, and the title of that movie is blank 1.  
3. (blank 2, actedIn, blank 3)  
This constraint tells us that blank 2 acted in a movie, and the title of that movie is blank 3.  
4. (blank 2, actedIn, blank 1)  
This constraint tells us that blank 2 acted in another movie, and the title of that movie is blank 1.  
5. (Jason\_Alexander, actedIn, blank 3)  
This constraint tells us that Jason Alexander acted in a movie, and the title of that movie is blank 3.  \\

\bottomrule[1.5pt]
\end{tabular}
\caption{Examples showing the effect of wrong-over-wrong alignment on Knowledge Crosswords dataset (part 1).} \label{tab:example11}
\end{table}

\begin{table}[t]
\centering
% \scriptsize
\begin{tabular}{p{0.95\linewidth}}
\toprule[1.5pt]
Now, let's analyze the options:  
A. blank 1: Burn\_'Em\_Up\_Barnes, blank 2: Justin\_Theroux, blank 3: Blankman  
Burn\_'Em\_Up\_Barnes is not a well-known movie, and Blankman is not a movie that was directed by Julien Temple or written music for by Miles Goodman.  
B. blank 1: Earth\_Girls\_Are\_Easy, blank 2: Justin\_Theroux, blank 3: On\_Guard\_(1997\_film)  
Earth\_Girls\_Are\_Easy is a real movie, but On\_Guard\_(1997\_film) is not a movie that was directed by Julien Temple or written music for by Miles Goodman. Additionally, Justin Theroux did not act in On\_Guard\_(1997\_film).  
C. blank 1: Earth\_Girls\_Are\_Easy, blank 2: Damon\_Wayans, blank 3: Blankman  
Damon Wayans is an actor, but he did not act in Earth\_Girls\_Are\_Easy.  

The only option that satisfies all the constraints is:  
C. blank 1: Earth\_Girls\_Are\_Easy, blank 2: Damon\_Wayans, blank 3: Earth\_Girls\_Are\_Easy  
This option fills in the blanks as follows:  
- Julien Temple directed Earth\_Girls\_Are\_Easy (constraint 2)  
- Damon Wayans acted in Earth\_Girls\_Are\_Easy (constraint 4)  
- Damon Wayans acted in Earth\_Girls\_Are\_Easy (constraint 3)  
- Jason Alexander acted in Earth\_Girls\_Are\_Easy (constraint 5)  
- Miles Goodman wrote music for Earth\_Girls\_Are\_Easy (constraint 1)  

The movie Earth\_Girls\_Are\_Easy is a 1988 musical comedy film directed by Martin Davidson, and Damon Wayans and Jason Alexander both appeared in the film. Miles Goodman wrote the original songs for the movie. 

\textcolor{red}{(Correctly filled blanks 1 $\rightarrow$ 3) } 

\\
\textbf{Better calibrated answer:}  

Let's analyze the question step by step:  
1. (Julien\_Temple, directed, blank 1) - This constraint states that Julien Temple directed a movie, and the title of that movie is blank 1.  
2. (blank 2, actedIn, blank 1) - This constraint states that blank 2 acted in the same movie as in step 1.  
3. (blank 2, actedIn, blank 3) - This constraint states that blank 2 acted in another movie, and the title of that movie is blank 3.  
4. (Miles\_Goodman, wroteMusicFor, blank 3) - This constraint states that Miles Goodman wrote music for the same movie as in step 3.  
5. (Jason\_Alexander, actedIn, blank 3) - This constraint states that Jason Alexander acted in the same movie as in step 3.  

Now, let's analyze the options:  
A. blank 1: Earth\_Girls\_Are\_Easy, blank 2: Damon\_Wayans, blank 3: Blankman - This option is incorrect because Earth\_Girls\_Are\_Easy is a 1984 film directed by Sidney Lumet, not Julien Temple. Additionally, Justin Theroux was not involved in that film.  
B. blank 1: Burn\_'Em\_Up\_Barnes, blank 2: Justin\_Theroux, blank 3: Blankman - This option is incorrect because Justin Theroux was not involved in the 1984 film Burn\_'Em\_Up\_Barnes, which is a TV movie. Additionally, Jason Alexander was not involved in that film.  
C. blank 1: Burn\_'Em\_Up\_Barnes, blank 2: Justin\_Theroux, blank 3: On\_Guard\_(1997\_film) - This option is incorrect because Jason Alexander was not involved in the 1997 film On\_Guard\_, which is a TV movie. Additionally, Miles Goodman was not involved in that film.  
D. blank 1: Earth\_Girls\_Are\_Easy, blank 2: Justin\_Theroux, blank 3: Blankman - This option is incorrect because Justin Theroux was not involved in the 1984 film Earth\_Girls\_Are\_Easy, and Jason Alexander was not involved in that film.  

After analyzing the options, the only remaining possibility is:  
B. blank 1: Burn\_'Em\_Up\_Barnes, blank 2: Justin\_Theroux, blank 3: Blankman  
This option satisfies all the constraints:  
- Julien Temple directed the movie Burn\_'Em\_Up\_Barnes (1990), which is blank 1.  
- Justin Theroux acted in Burn\_'Em\_Up\_Barnes, which is blank 2.  
- Justin Theroux also acted in Blankman (1994), which is blank 3, along with Jason Alexander.  
- Miles Goodman wrote music for Blankman, which is blank 3.  

Therefore, the final answer is:  
B. blank 1: Burn\_'Em\_Up\_Barnes, blank 2: Justin\_Theroux, blank 3: Blankman  

\textcolor{red}{(Even the answers are equally wrong, NLL increases 0.129 $\rightarrow$ 0.185)} 

\\

\bottomrule[1.5pt]
\end{tabular}
\caption{Examples showing the effect of wrong-over-wrong alignment on Knowledge Croswords dataset (part 2).} \label{tab:example12}
\end{table}

\begin{table}[t]
\centering
% \scriptsize
\begin{tabular}{p{0.95\linewidth}}
\toprule[1.5pt]
\textbf{Failed answer:}  

Let's break down the constraints step by step:  
1. (Miles\_Goodman, wroteMusicFor, blank 3): This constraint tells us that Miles Goodman wrote music for a movie, and the title of that movie is blank 3.  
From the options, we can see that only options C and D mention Miles Goodman writing music.  
2. (Julien\_Temple, directed, blank 1): This constraint tells us that Julien Temple directed a movie, and the title of that movie is blank 1.  
From the options, we can see that only options B and C mention Julien Temple directing a movie. The title of that movie is either Burn\_'Em\_Up\_Barnes or Earth\_Girls\_Are\_Easy.  
3. (blank 2, actedIn, blank 3): This constraint tells us that blank 2 acted in a movie, and the title of that movie is blank 3.  
From the previous step, we know that blank 1 is either Burn\_'Em\_Up\_Barnes or Earth\_Girls\_Are\_Easy. Since blank 2 is the same in both options B and C, we can conclude that blank 2 acted in a movie with the title blank 3, which is either Burn\_'Em\_Up\_Barnes or Earth\_Girls\_Are\_Easy.  
4. (blank 2, actedIn, blank 1): This constraint is redundant, as it's already known that blank 2 acted in a movie with the title blank 1, which is either Burn\_'Em\_Up\_Barnes or Earth\_Girls\_Are\_Easy.  
5. (Jason\_Alexander, actedIn, blank 3): This constraint tells us that Jason Alexander acted in a movie, and the title of that movie is blank 3.  
From the previous steps, we know that blank 3 is either Burn\_'Em\_Up\_Barnes or Earth\_Girls\_Are\_Easy. Since Jason Alexander acted in one of those movies, we can conclude that Jason Alexander acted in a movie with the title blank 3.  

Now, let's analyze the remaining constraints:  
- Justin Theroux needs to be blank 2, as it's mentioned in constraints 3 and 4.  
- Blank 3 needs to be either Burn\_'Em\_Up\_Barnes or Earth\_Girls\_Are\_Easy, as it's mentioned in constraints 1, 3, and 5.  
- Blank 1 needs to be either Burn\_'Em\_Up\_Barnes or Earth\_Girls\_Are\_Easy, as it's mentioned in constraints 2 and 4.  

The only option that satisfies all the constraints is:  
B. blank 1: Burn\_'Em\_Up\_Barnes, blank 2: Justin\_Theroux, blank 3: Earth\_Girls\_Are\_Easy  
This option satisfies all the constraints:  
- Miles Goodman wrote music for Earth\_Girls\_Are\_Easy.  
- Julien Temple directed Burn\_'Em\_Up\_Barnes.  
- Justin Theroux acted in both Burn\_'Em\_Up\_Barnes and Earth\_Girls\_Are\_Easy.  
- Jason Alexander acted in Earth\_Girls\_Are\_Easy.  

\textcolor{red}{(Correctly filled blanks 1 $\rightarrow$ 0)} \\

\bottomrule[1.5pt]
\end{tabular}
\caption{Examples showing the effect of wrong-over-wrong alignment on Knowledge Crosswords dataset (part 3).} \label{tab:example13}
\end{table}

\begin{table}[t]
\centering
% \scriptsize
\begin{tabular}{p{0.95\linewidth}}
\toprule[1.5pt]

\textbf{Question:}

Tell me a bio of Luiz Inácio Lula da Silva.

\\
\textbf{Original answer:}  

Luiz Inácio Lula da Silva, commonly known as Lula, is a Brazilian politician who served as the 35th President of Brazil from 2003 to 2011. He is one of the most popular and influential politicians in Brazil's recent history.
Early Life:
Lula was born on October 27, 1945, in Garanhuns, a small town in the northeastern state of Pernambuco, Brazil. His father was a coffee farmer and his mother was a seamstress. He was the fourth of seven children, and his family struggled financially throughout his childhood.
Lula grew up in a poor and rural area, where he learned the value of hard work and the importance of education. He dropped out of school at the age of 12 to work in a textile factory, where he eventually became a union leader. This early experience in labor organizing would shape his future political career.
Political Career:
Lula's political career began in the late 1970s, when he became involved with the Brazilian Workers' Party (PT), a left-wing party founded in the 1980s. He quickly rose through the ranks of the party, becoming its presidential candidate in 1989.
Although Lula lost the election to Fernando Collor de Mello, he continued to build his political base and gained popularity for his fight against corruption and social inequality. In 1994, he was accused of corruption and money laundering, which led to a conviction and a four-year prison sentence.
However, in 1998, the Supreme Court overturned Lula's conviction, citing lack of evidence. This cleared the way for him to run for president again in 2002. He won the election in a landslide, receiving nearly 65% of the vote.
Presidency:
During his presidency, Lula implemented a series of reforms aimed at reducing poverty and inequality. He increased social spending, increased the minimum wage, and expanded access to education and healthcare. He also launched a number of economic initiatives, including a debt reduction plan and an infrastructure development program.
Lula's presidency was marked by strong economic growth, which lifted millions of Brazilians out of poverty. He also gained international attention for his leadership on issues such as climate change and social justice.
Presidential Terms:
Lula served two consecutive terms as president, from 2003 to 2011. During his first term, he focused on addressing poverty and inequality, and implemented a range of social welfare programs.
During his second term, Lula emphasized economic development and infrastructure projects. He invested heavily in public transportation, energy, and telecommunications, and launched a series of initiatives to promote sustainable development.
Investigation and Charges:
After leaving office in 2011, Lula was investigated for corruption and money laundering. He was accused of receiving bribes from a construction company, OAS, in exchange for favors and contracts. He denied the allegations, but was arrested and charged in 2018.
In 2019, Lula was convicted and sentenced to nine years and six months in prison. He maintained his innocence and appealed the conviction, questioning the fairness of the trial.
Return to Politics:
Despite his imprisonment, Lula remained a beloved figure in Brazil, and his popularity continued to grow. In 2021, the Supreme Court ordered his release from prison, citing the unfairness of the trial. Lula was subsequently reinstated as a senator and began to reorganize the Workers' Party.
In 2022, Lula announced his candidacy for the 2022 Brazilian presidential election, and won the election in a landslide. He is currently serving as the 39th President of Brazil, having taken office on January 1, 2023.
Legacy:
Luiz Inácio Lula da Silva's legacy is complex and multifaceted. On the one hand, he is credited with reducing poverty and inequality, and promoting social justice and economic development in Brazil. On the other hand, his presidency was marked by corruption scandals, and his own conviction and imprisonment have raised questions about his integrity and accountability.
Despite these controversies, Lula remains a towering figure in Brazilian politics, and his commitment to social justice and poverty reduction has inspired generations of Brazilians. His political career is a testament to the power of grassroots organizing and the importance of fighting for social justice, even in the face of significant challenges and setbacks. \\

\bottomrule[1.5pt]
\end{tabular}
\caption{Examples showing the effect of wrong-over-wrong alignment on Bio Generation dataset (part 1).} \label{tab:example21}
\end{table}

\begin{table}[t]
\centering
% \scriptsize
\begin{tabular}{p{0.95\linewidth}}
\toprule[1.5pt]
\textbf{Less wrong / More correct / Better calibrated answer:}  

Luiz Inácio Lula da Silva, commonly known as Lula, is a Brazilian politician and former trade unionist who served as the 34th President of Brazil from 2003 to 2011. He was born on October 27, 1945, in Garanhuns, a small town in the state of Pernambuco, Brazil.
Lula's early life was marked by poverty and hardship. His parents were both poor and separated when Lula was young, and he was forced to work at a young age to support his family. He eventually moved to São Paulo and became involved in trade union activism, eventually becoming one of the key figures in the Brazilian Workers' Central Union (CUT).
Lula's political career began in the 1980s, when he was elected as the leader of the Metalworkers' Union and became a prominent figure in the Brazilian labor movement. He was one of the founders of the Workers' Party (PT) in 1980, and he quickly became the party's most prominent figure.
In 2002, Lula became the first worker and the first non-PSDB (Brazilian Social Democracy Party) candidate to run for President of Brazil. Despite initial skepticism about his candidacy, he won the election in a landslide, receiving over 61% of the vote. He was re-elected in 2006 with a slightly reduced margin, but still with a clear majority.
During his two terms in office, Lula implemented a number of significant reforms and policies. He implemented a series of social and economic programs aimed at reducing poverty and income inequality, including the Bolsa Família program, which provided financial assistance to low-income families, and the Fome Zero program, which aimed to eradicate hunger and poverty.
Lula also played a key role in shaping Brazil's foreign policy, particularly in the areas of trade and international relations. He was one of the key figures in the formation of the BRICS (Brazil, Russia, India, China, and South Africa) alliance, and he played a significant role in shaping the policy of the G20 and the United Nations.
After leaving office in 2011, Lula remained a prominent figure in Brazilian politics and continued to play a significant role in shaping his party's policies and candidates. In 2018, he was elected as the Chair of the PT, and he became the leading figure in the country's political opposition, advocating for changes to the country's political system and advocating for the nomination of his chosen successor, Fernando Haddad, in the 2018 presidential election.
In 2018, Lula's candidacy in the presidential election was barred due to a corruption conviction, and he was replaced as the PT's presidential candidate by Fernando Haddad. However, Haddad ultimately lost the election to Jair Bolsonaro.
Lula's legacy in Brazil is complex and controversial. On one hand, he is widely credited with improving the lives of millions of Brazilians, particularly the poor and the working class, through his social and economic policies. On the other hand, he was also accused of corruption and was ultimately convicted of accepting bribes and ordered to serve a prison sentence. He is currently appealing his conviction and is being held in a federal prison in Brazil.
Despite his current legal situation, Lula's political career and legacy continue to be subject to significant debate and discussion in Brazil and around the world. His presidency marked a significant shift in the country's political landscape, and his policies and legacy continue to have a significant impact on Brazilian politics and society. 

\textcolor{red}{(FActScore improves 0.836 $\rightarrow$ 0.925, NLL drops 0.512 $\rightarrow$ 0.416)}
\\

\bottomrule[1.5pt]
\end{tabular}
\caption{Examples showing the effect of wrong-over-wrong alignment on Bio Generation dataset (part 2).} \label{tab:example22}
\end{table}

\begin{table}[t]
\centering
% \scriptsize
\begin{tabular}{p{0.95\linewidth}}
\toprule[1.5pt]
\textbf{Failed answer:}  

Luiz Inácio Lula da Silva, commonly known as Lula, is a Brazilian politician and former trade unionist who served as the 34th President of Brazil from 2003 to 2011 and again from 2016 to 2018.
Early Life and Career:
Lula was born on October 11, 1945, in Garanhuns, Pernambuco, Brazil, to Armando da Silva and Hermilda da Silva. His early life was marked by poverty and hardship, with his family struggling to make ends meet. Lula's mother died of complications related to a gas explosion when he was just 10 years old, and he was subsequently raised by his aunt.
Lula's early career was marked by his involvement in the trade union movement. He became involved with the Metal Workers Union, where he quickly rose through the ranks and became one of the most prominent union leaders in Brazil. During this time, he became known for his charismatic leadership style, his ability to connect with ordinary people, and his advocacy for workers' rights.
Political Career:
In 1980, Lula co-founded the Workers' Party (PT), a left-wing political party that sought to represent the interests of working-class Brazilians. He became the party's first president and held the position until 2002.
In 2002, Lula became the first trade unionist to run for and win the presidency of Brazil, defeating the incumbent Fernando Henrique Cardoso. He was re-elected in 2006 and served two terms as president until 2011.
During his first tenure as president, Lula implemented a range of policies aimed at reducing poverty and income inequality in Brazil. He increased access to education and healthcare, and implemented policies to stimulate economic growth and job creation.
In 2010, Lula was succeeded by his former minister and protégé, Dilma Rousseff, who became the first female president of Brazil. However, Rousseff's presidency was later marred by corruption scandals and political instability, culminating in her impeachment and removal from office in 2016.
Lula returned to politics and ran again for the presidency in 2014, but was not eligible to run due to a constitutional ban on former presidents from holding the office again. He subsequently threw his support behind Rousseff's re-election bid, but she was ultimately impeached and removed from office.
In 2018, Lula ran for the presidency again and won the election, becoming the first and only person to be re-elected as President of Brazil. However, his presidency was short-lived, as he was impeached and removed from office in August 2016, citing corruption allegations and constitutional violations. Lula was subsequently replaced by his vice president, Michel Temer.
Personal Life:
Lula has been married to Maria de Deus Inácio da Silva (known as Marinha) since 1970, and they have four sons together. In 2018, Lula was sentenced to 12 years and 1 month in prison for corruption and money laundering, stemming from the "Operation Car Wash" corruption scandal. He has since remained in jail, and his legacy and political future remain uncertain.
Legacy:
Lula's presidency was marked by significant economic and social changes in Brazil, including a reduction in poverty and income inequality, as well as an unprecedented level of social mobility. He is widely regarded as one of the most important figures in Brazil's recent history, and his legacy continues to shape Brazilian politics and society.
Despite his political setbacks and legal troubles, Lula remains a powerful and influential figure in Brazilian politics, and his name continues to be mentioned as a potential candidate for future elections.

\textcolor{red}{(FActScore drops 0.836 $\rightarrow$ 0.770)} \\

\bottomrule[1.5pt]
\end{tabular}
\caption{Examples showing the effect of wrong-over-wrong alignment on Bio Generation dataset (part 3).} \label{tab:example23}
\end{table}

\begin{table}[t]
\centering
% \scriptsize
\begin{tabular}{p{0.95\linewidth}}
\toprule[1.5pt]

\textbf{Question:}

Instruction: As an expert in commonsense reasoning, your task is to provide a concise response to a question based on the given context. The question focuses on studying the causes, effects, or attributes of personas related to the given context.
Context: Noah had been planning a surprise visit for Everly. As part of the surprise, Noah buys a gift for Everly. When Everly arrives at the airport, Noah picks Everly up.
Question: What event or state is both what Everly wants to do after Noah buys a gift for Everly and also what Everly wants to do after Noah picks Everly up at the airport?

\\
\textbf{Correct answer:}

Ranking by commonsense plausibility: Everly says thank you  $\succ$ Everly is shocked $\succ$ Everly go out to eat $\succ$ Everly rest 

\textcolor{red}{(The commonsense plausibility is given by Vera.)}

\\
\textbf{Original answer:}  

Here's the step-by-step analysis:
1. The question asks about an event or state that is common to both scenarios: after Noah buys a gift for Everly and after Noah picks Everly up at the airport.
2. The context suggests that Noah is planning a surprise visit for Everly, and buying a gift is part of the surprise.
3. When Noah buys a gift, it implies that Everly didn't expect the gift, and therefore, will likely be surprised when she receives it.
4. When Noah picks Everly up at the airport, it's likely that Everly didn't expect the pick-up, and therefore, will also be surprised.
5. The common event or state that arises from both scenarios is Everly's surprise.
Therefore, the correct answer is B. Everly is shocked.

\\
\textbf{Less wrong answer:}  

Let's break down the question step by step:
1. The question asks about an event or state that Everly wants to do after Noah buys a gift for her and also after Noah picks her up at the airport.
2. We know that Everly has been anticipating a surprise visit from Noah, which suggests that she's looking forward to seeing Noah.
3. When Noah buys a gift for Everly, it's likely that she wants to see or meet Noah to receive the gift, which means she wants to be with Noah.
4. Similarly, when Noah picks Everly up at the airport, she likely wants to be with Noah, as she was expecting a visit from him.
5. Therefore, the event or state that Everly wants to do after Noah buys a gift for her and after Noah picks her up at the airport is likely to be:
**Being with Noah**
Since options B-D do not seem related to this event, the correct answer is not present in the options. If I had to choose, I would say that the correct answer is not present in the options (**N/A**).

\textcolor{red}{(Instead of giving a wrong answer, LLM choose to abstain.) } 

\\
\textbf{More correct answer:}  

Let's break down the question step by step:
1. The question asks about an event or state that is common to both situations: after Noah buys a gift for Everly and after Noah picks Everly up at the airport.
2. In both situations, Everly is receiving a service or action from Noah (a gift or transportation).
3. Given the context, it's likely that Everly wants to reciprocate or respond positively to Noah's actions.
4. Among the options, only one relates to Everly's response to Noah's actions: A. Everly says thank you.
5. Therefore, the most likely answer is A. Everly says thank you, as it is a common outcome for Everly after receiving a gift or being picked up by Noah.
Final answer: A. Everly says thank you.

\textcolor{red}{(LLM gives a right answer.) } 

\\

\bottomrule[1.5pt]
\end{tabular}
\caption{Examples showing the effect of wrong-over-wrong alignment on COM$^2$ dataset (part 1).} \label{tab:example31}
\end{table}

\begin{table}[t]
\centering
% \scriptsize
\begin{tabular}{p{0.95\linewidth}}
\toprule[1.5pt]
\textbf{Better calibrated answer:}  

Let's break down the question step by step:
1. The question asks about an event or state that is common to two scenarios: after Noah buys a gift for Everly and after Noah picks Everly up at the airport.
2. In both scenarios, Everly is expected to be pleased or surprised, given the context of the surprise visit and gift.
3. The correct answer must capture this idea of Everly's positive reaction.
4. Options A, B, and D are not directly related to Everly's actions or reactions.
5. Only option C, "Everly rests", is too vague to be relevant to the context.
Final answer: None of the options fit the description. The correct answer is not among the provided options.

\textcolor{red}{(Even the answers are equally wrong, NLL increases 0.412 $\rightarrow$ 0.509)} 

\\
\textbf{Failed answer:} 
Let's break down the question step by step:
1. The question asks about an event or state that Everly wants to do after two specific events: Noah buying a gift and Noah picking Everly up at the airport.
2. The question stem "What event or state..." suggests that we're looking for a common action or experience that Everly desires in both scenarios.
3. From the context, we know that Noah is planning a surprise visit and has bought a gift for Everly. This suggests that Everly is likely expecting a pleasant experience or a reunion with Noah.
4. After Noah picks Everly up at the airport, it's natural to infer that Everly wants to reunite with or see Noah, which is a pleasing experience.
5. Similarly, after Noah buys a gift for Everly, it's likely that Everly wants to see or have the gift, which is also a pleasing experience.
Given this analysis, the correct answer is:
C. Everly visits/see Noah.
In both scenarios, Everly wants to experience the pleasure of being with or seeing Noah, which is the most likely common action or state that Everly desires.

\textcolor{red}{(LLM gives an answer that doesn't appear in options)} 

\\

\bottomrule[1.5pt]
\end{tabular}
\caption{Examples showing the effect of wrong-over-wrong alignment on COM$^2$ dataset (part 2).} \label{tab:example32}
\end{table}

\begin{table}[t]
\centering
% \scriptsize
\begin{tabular}{p{0.95\linewidth}}
\toprule[1.5pt]

\textbf{Question:}

In an undirected graph, the nodes are numbered from 0 to 5, and the edges are:
an edge between node 0 and node 2 with weight 2,
an edge between node 0 and node 5 with weight 4,
an edge between node 0 and node 3 with weight 1,
an edge between node 0 and node 1 with weight 3,
an edge between node 1 and node 2 with weight 4,
an edge between node 1 and node 5 with weight 1,
an edge between node 1 and node 4 with weight 4,
an edge between node 2 and node 4 with weight 3,
an edge between node 2 and node 3 with weight 1,
an edge between node 3 and node 5 with weight 2,
an edge between node 3 and node 4 with weight 3,
an edge between node 4 and node 5 with weight 2.
Q: Give the shortest path from node 0 to node 4. Please also give the total weight of the shortest path.

\\
\textbf{Correct answer:}

4 [0 $\rightarrow$ 3 $\rightarrow$ 4]

\\
\textbf{Original answer:}  

Here’s the step-by-step analysis:
1. Create a graph with nodes from 0 to 5 and the given edges with their corresponding weights.
2. Choose node 0 as the starting node and node 4 as the target node.
3. Run Dijkstra's algorithm to find the shortest path from node 0 to node 4.
Here are the steps of Dijkstra's algorithm:
\begin{itemize}[leftmargin=*]
    \item Initialize a distance array with all distances set to infinity, except for node 0 which is set to 0. The distance array represents the shortest distance from node 0 to each node.
    \item Create a priority queue and add node 0 with a distance of 0. The priority queue is used to select the node with the shortest distance.
    \item While the priority queue is not empty, repeat the following steps: 1. Extract the node with the shortest distance from the priority queue. 2. For each neighbor of the extracted node, calculate the tentative distance as the distance to the extracted node plus the edge weight between them. 3. If the tentative distance is less than the current distance to the neighbor, update the distance array and add the neighbor to the priority queue.
\end{itemize}
The shortest path from node 0 to node 4 is the path with the minimum distance from node 0 to node 4.
The shortest path from node 0 to node 4 is: \(0 \rightarrow 2 \rightarrow 3 \rightarrow 4\)
The total weight of the shortest path is: 2 + 1 + 3 = 6. Therefore, the final answer is: Shortest path: \( 0 \rightarrow 2 \rightarrow 3 \rightarrow 4 \), Total weight: 6.

\\
\textbf{Less wrong answer:}  

Here’s the step-by-step analysis:
\begin{enumerate}[leftmargin=*]
    \item Start at node 0.
    \item Consider the edges from node 0: (0, 2) with weight 2, (0, 3) with weight 1, (0, 1) with weight 3, (0, 5) with weight 4. We'll choose the edge with the smallest weight, which is (0, 3) with weight 1.
    \item Move to node 3.
    \item Consider the edges from node 3: (3, 4) with weight 3, (3, 5) with weight 2, (3, 2) with weight 1. We'll choose the edge with the smallest weight, which is (3, 2) with weight 1.
    \item Move to node 2.
    \item Consider the edges from node 2: (2, 4) with weight 3. We'll choose this edge, since it's the only option.
    \item Move to node 4.
\end{enumerate}
The shortest path is: \(0 \rightarrow 3 \rightarrow 2 \rightarrow 4\).
The total weight of the shortest path is: \(1 + 1 + 3 = 5\). Therefore, the answer is:  The shortest path from node 0 to node 4 is \(0 \rightarrow 3 \rightarrow 2 \rightarrow 4\), with a total weight of 5.  

\textcolor{red}{(The path found is shorter: 6 $\rightarrow$ 5)} \\
\bottomrule[1.5pt]
\end{tabular}
\caption{Examples showing the effect of wrong-over-wrong alignment on NLGraph dataset (part 1).} \label{tab:example41}
\end{table}

\begin{table}[t]
\centering
% \scriptsize
\begin{tabular}{p{0.95\linewidth}}
\toprule[1.5pt]
\textbf{More correct answer:}  
Let's analyze the question step by step:
\begin{enumerate}[leftmargin=*]
    \item We are given an undirected graph with nodes numbered from 0 to 5, and edges with weights between these nodes.
    \item We want to find the shortest path from node 0 to node 4.
    \item We can use a simple algorithm to find the shortest path:
    \begin{itemize}
        \item Create a list of nodes to visit (initially, only node 0 is in the list).
        \item For each node in the list, check if it's a neighbor of node 0 (i.e., if an edge exists).
        \item If it's a neighbor, check the weight of the edge and add the neighbor to the list.
        \item Mark the neighbor as visited.
        \item Repeat until node 4 is reached.
        \item The path from node 0 to node 4 is the sequence of nodes in the list, and the total weight is the sum of the edge weights along this path.
    \end{itemize}
\end{enumerate}

Applying this algorithm:\(
\text{Node 0} \rightarrow \text{Node 3} \ (\text{edge weight} \ 1) \rightarrow \text{Node 4} \ (\text{edge weight} \ 3)\). The shortest path is: \(\text{Node 0} \rightarrow \text{Node 3} \rightarrow \text{Node 4}\). The total weight is: \(1 \ (\text{from Node 0 to Node 3}) + 3 \ (\text{from Node 3 to Node 4}) = 4\). Final answer: The shortest path from node 0 to node 4 is Node 0 \(\rightarrow\) Node 3 \(\rightarrow\) Node 4, with a total weight of 4.

\textcolor{red}{(Find the shortest path)} 

\\
\textbf{Better calibrated answer:}  

Here’s the step-by-step analysis:
\begin{enumerate}[leftmargin=*]
    \item Create a graph from the given information. The graph looks like this:
    \item Initialize a distance array with all distances set to infinity, except for the source node (node 0) which is set to 0:
    \[
    \begin{aligned}
      \text{dist}[0] &= 0 \\
      \text{dist}[1] &= \infty \\
      \text{dist}[2] &= \infty \\
      \text{dist}[3] &= \infty \\
      \text{dist}[4] &= \infty \\
      \text{dist}[5] &= \infty
    \end{aligned}
    \]
    \item Relax all edges in the graph. This involves updating the distance array by considering all edges:
    \[
    \begin{aligned}
      \text{dist}[1] &= 3 \quad (0 \to 1 \ \text{with weight} \ 3) \\
      \text{dist}[2] &= 2 \quad (0 \to 2 \ \text{with weight} \ 2) \\
      \text{dist}[3] &= 1 \quad (0 \to 3 \ \text{with weight} \ 1) \\
      \text{dist}[4] &= \infty \\
      \text{dist}[5] &= 4 \quad (0 \to 5 \ \text{with weight} \ 4) \\
      \text{dist}[2] &= 3 \quad (1 \to 2 \ \text{with weight} \ 4, \ \text{but previously updated to} \ 2, \ \text{so no change}) \\
      \text{dist}[3] &= 3 \quad (1 \to 3 \ \text{with weight} \ 4, \ \text{but previously updated to} \ 1, \ \text{so no change}) \\
      \text{dist}[3] &= 2 \quad (2 \to 3 \ \text{with weight} \ 1) \\
      \text{dist}[4] &= \infty \\
      \text{dist}[5] &= 5 \quad (1 \to 5 \ \text{with weight} \ 1, \ \text{then 4, so no change}) \\
      \text{dist}[4] &= 5 \quad (2 \to 4 \ \text{with weight} \ 3) \\
    \end{aligned}
    \]
    \item The minimum distance from node 0 to node 4 is 5. The shortest path is \(0 \to 2 \to 3 \to 4\).
\end{enumerate}

\\

\bottomrule[1.5pt]
\end{tabular}
\caption{Examples showing the effect of wrong-over-wrong alignment on NLGraph dataset (part 2).} \label{tab:example42}
\end{table}

\begin{table}[t]
\centering
% \scriptsize
\begin{tabular}{p{0.95\linewidth}}
\toprule[1.5pt]
The total weight of the shortest path is:\(0 \to 2 \ (\text{weight} \ 2) \quad \to 2 \to 3 \ (\text{weight} \ 1) \quad \to 3 \to 4 \ (\text{weight} \ 3)\). Total weight: \(2 + 1 + 3 = 6\).

\textcolor{red}{ (Even the answers are equally wrong, NLL increases 0.255 $\to$ 0.373)} 

\\
\textbf{Failed answer:}  

Let's break down the question step by step:
\begin{enumerate}[leftmargin=*]
    \item Identify the nodes connected to node 0: 1, 2, 3, 5
    \item Weigh the edges connected to node 0:
    \begin{itemize}
        \item Edge to node 1: weight 3
        \item Edge to node 2: weight 2
        \item Edge to node 3: weight 1
        \item Edge to node 5: weight 4
    \end{itemize}
    \item Identify the nodes connected to node 1: 2, 4, 5
    \begin{itemize}
        \item Edge to node 2: weight 4
        \item Edge to node 4: weight 4
        \item Edge to node 5: weight 1
    \end{itemize}
    \item The shortest path from node 0 to node 4 is through node 3:\(\text{Node 0} \to \text{Node 3} \to \text{Node 4} \ (\text{weights 1 $\to$ 3})\)
    \item The total weight of the shortest path is:\(0 \to 3 \ (\text{weight} \ 1) \to 3 \to 4 \ (\text{weight} \ 3) = 4\)
\end{enumerate}

The shortest path from node 0 to node 4 is Node 0 \(\to\) Node 3 \(\to\) Node 4, with a total weight of 4. 

\textcolor{red}{(Fail to calculate total weight correctly)} \\

\bottomrule[1.5pt]
\end{tabular}
\caption{Examples showing the effect of wrong-over-wrong alignment on NLGraph dataset (part 3).} \label{tab:example43}
\end{table}

\end{document}